\documentclass[sigconf]{acmart}
\usepackage{multirow}
\usepackage{caption}
\usepackage{subcaption}
\usepackage{makecell}
\usepackage{bm}
\usepackage{soul, color, xcolor}

\AtBeginDocument{%
  }

\setcopyright{acmcopyright}
\copyrightyear{2023}
\acmYear{2023}
\acmDOI{XXXXXXX.XXXXXXX}

\acmPrice{15.00}
\acmISBN{978-1-4503-XXXX-X/18/06}

\begin{document}

\title{Advancing Incremental Few-shot Semantic Segmentation via Semantic-guided Relation Alignment and Adaptation}

%
%
%
\author{Yuan Zhou}
\affiliation{%
 \institution{Hefei University of Technology}
 \city{Hefei}
 \country{China}}
 \email{2018110971@mail.hfut.edu.cn}

\author{Xin Chen}
\affiliation{%
  \institution{Huawei Inc.}
  \city{Shenzhen}
  \country{China}}
\email{chenxin061@gmail.com}

\author{Yanrong Guo}
\affiliation{%
 \institution{Hefei University of Technology}
 \city{Hefei}
 \country{China}}
 \email{yrguo@hfut.edu.cn}

\author{Shijie Hao}
\affiliation{%
 \institution{Hefei University of Technology}
 \city{Hefei}
 \country{China}}
 \email{hfut.hsj@gmail.com}

\author{Richang Hong}
\affiliation{%
 \institution{Hefei University of Technology}
 \city{Hefei}
 \country{China}}
 \email{hongrc.hfut@gmail.com}

\author{Qi Tian}
\affiliation{%
  \institution{Huawei Inc.}
  \city{Shenzheng}
  \country{China}}
\email{tian.qi1@huawei.com}


\begin{abstract}
Incremental few-shot semantic segmentation (IFSS) aims to incrementally extend a semantic segmentation model to novel classes according to only a few pixel-level annotated data, while preserving its segmentation capability on previously learned base categories. This task faces a severe semantic-aliasing issue between base and novel classes due to data imbalance, which makes segmentation results unsatisfactory. To alleviate this issue, we propose the Semantic-guided Relation Alignment and Adaptation (SRAA) method that fully considers the guidance of prior semantic information. Specifically, we first conduct Semantic Relation Alignment (SRA) in the base step, so as to semantically align base class representations to their semantics. As a result, the embeddings of base classes are constrained to have relatively low semantic correlations to categories that are different from them. Afterwards, based on the semantically aligned base categories, Semantic-Guided Adaptation (SGA) is employed during the incremental learning stage. It aims to ensure affinities between visual and semantic embeddings of encountered novel categories, thereby making the feature representations be consistent with their semantic information. In this way, the semantic-aliasing issue can be suppressed. We evaluate our model on the PASCAL VOC 2012 and the COCO dataset. The experimental results on both these two datasets exhibit its competitive performance, which demonstrates the superiority of our method.
\end{abstract}

  


\received{10 March 2023}

\maketitle
\section{Introduction}
In recent years, semantic segmentation has achieved impressive performance by using deep neural networks. However, conventional semantic segmentation models generally have a fixed output space. Therefore, when encountering new categories, they need to be re-trained from scratch to update their segmentation capability. Moreover, these models require large-scale pixel-level labeled data, which are expensive and laborious to obtain. These issues limit their applicability in open-ended real-world scenarios. In this context, Cermelli et al. \cite{cermelli2021prototype} proposed the Incremental Few-shot Semantic Segmentation (IFSS) task. It aims to effectively extend a semantic segmentation model to new categories using a few labeled samples, while maintaining its segmentation capability on previously learned old ones. In this way, the extendibility and flexibility of the model can be improved, which is critical for many real-world applications, such as autonomous driving and human-machine interaction. \par

\begin{figure}[t!]
\centering
\includegraphics[height=6.6cm]{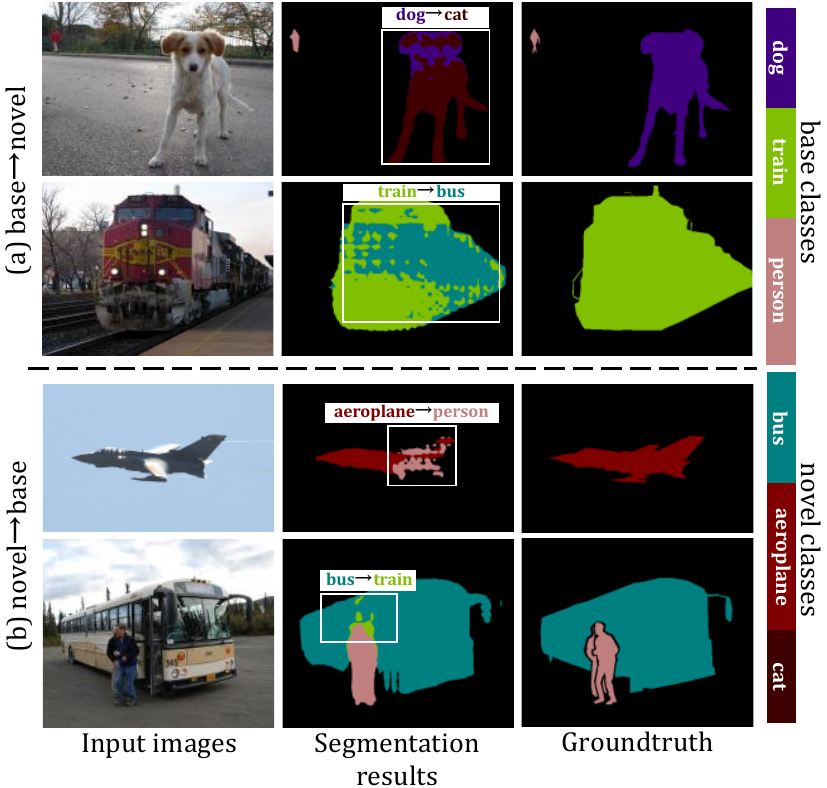}
\caption{The typical examples for the semantic-aliasing issue in IFSS, which are obtained from the PASCAL VOC 2012 dataset on the 1-shot task. ``A$\rightarrow$B'' indicates that regions belonging to A are incorrectly segmented as B.}
\label{fig:1}
\end{figure}

More specifically, in the IFSS task, a base set with relatively more training samples is first provided to initialize the learnable parameters of a semantic segmentation model. Then, a few pixel-level annotated training samples of novel categories are given, helping incrementally expand the segmentation capability of the model to the encountered novel ones. However, the IFSS model is prone to fall into the semantic-aliasing issue due to data imbalance between base and novel classes. As shown in Figure~\ref{fig:1}, the semantic confusion between the base class ``dog'' and the encountered novel category ``cat'' misleads the model to draw the incorrect segmentation results, making the model performance unsatisfactory. Recently, semantic information has been successfully introduced in the few-shot classification task \cite{li2020boosting,zhang2021prototype,xu2022generating,yang2023semantic}, aiming to make feature embeddings more representative, e.g., GloVe \cite{pennington2014glove} or word2vec \cite{mikolov2013efficient} was employed in \cite{li2020boosting,zhang2021prototype} to provide prior semantic information while \cite{xu2022generating,yang2023semantic} additionally considered the semantic guidance of CLIP \cite{radford2021learning}. \par

Inspired by these methods, we propose to suppress the semantic-aliasing issue in IFSS by fully considering the guidance of visual semantics. Therefore, we propose the Semantic-guided Relation Alignment and Adaptation (SRAA) method in this paper, which is shown in Figure \ref{fig:2}. On one hand, we propose to conduct Semantic Relation Alignment (SRA) in the base step, aiming to semantically align base class representations in latent semantic space. Therefore, the embeddings of base classes are constrained to have relatively low semantic correlations to categories that are different from them. Moreover, the cross-entropy loss is employed during this process to measure discrepancy between segmentation results and groundtruth label maps. As a result, the model is trained to segment base classes, while being aware of their semantic information. Based on the aligned base classes, Semantic-Guided Adaptation (SGA) is employed to incrementally adapt the model to novel classes. It aims to ensure affinities between visual and semantic embeddings of novel categories, thereby making the feature representations be consistent with their semantic information. By considering the semantic information of both the base and the novel classes, the semantic-aliasing issue can be alleviated. We evaluate our method on the public semantic segmentation datasets PASCAL VOC 2012 and COCO, following the cross validation used in \cite{cermelli2021prototype}. On both these datasets, our method presents competitive performance. 

All-in-all, the contributions of this paper can be summarized below:
\begin{itemize}
\item In this paper, we propose to suppress the semantic-aliasing issue in IFSS by fully considering the guidance of semantic information, thereby making segmentation results more accurate. To realize this goal, we accordingly propose the Semantic-guided Relation Alignment and Adaptation (SRAA) method.
\item We propose to conduct Semantic Relation Alignment (SRA) in the base step, aiming to semantically align the representations of base categories. Therefore, the base class embeddings are guided to have relatively low semantic correlations to categories that are different from them.
\item Based on the aligned base classes, we propose to conduct Semantic-Guided Adaptation (SGA) during the incremental learning stage, guiding the embeddings of novel classes to be consistent with their semantic information. In this way, the semantic aliasing between the base and the encountered novel categories can be alleviated. 
\end{itemize}

\section{Related Work}
In this section, we review methods that are relevant to our research. We first briefly introduce typical methods of semantic segmentation, few-shot learning, and incremental learning in section 2.1, section 2.2, and section 2.3. Then, we review related incremental few-shot semantic segmentation methods in section 2.4, and introduce their differences to our work.

\subsection{Semantic Segmentation}
Semantic segmentation, a pixel-level image recognition technique, has achieved remarkable progress in recent years with development of deep learning. \cite{long2015fully} is a typical deep-learning-based semantic segmentation method that uses the fully convolutional layer to realize efficient end-to-end dense predications for input images. Inspired by \cite{long2015fully}, many semantic segmentation models have been proposed. Zhao et al. \cite{zhao2017pyramid} further introduced the pyramid pooling module, aiming to fully aggregate global context information of visual scenes. Chen et al. \cite{chen2017deeplab,chen2017rethinking} proposed to aggregate multi-scale context information using the atrous convolution, so as to make segmentation results more accurate. Based on \cite{long2015fully,chen2017deeplab,chen2017rethinking}, Zhang et al. \cite{zhang2018context} learned an inherent dictionary to aggregate semantic context information of a whole dataset, which help the model understand visual scenes in a more global way. Huang et al. \cite{huang2019ccnet} enhanced a semantic segmentation model with the proposed criss-cross attention layer. Therefore, sufficient context information is aggregated for each pixel, while the model is maintained with high efficiency. Recently, the methods \cite{xie2021segformer,zheng2021rethinking,strudel2021segmenter} have successfully built semantic segmentation models upon the transformer \cite{vaswani2017attention,dosovitskiy2020image}, thereby further boosting visual representations of input images.

\subsection{Few-shot Learning}
Few-shot learning aims to quickly transfer models to novel unseen categories according to only one or a few training instances, thereby reducing expenses cost on data preparation. Currently, few-shot learning methods are mainly based on metric learning, aiming at learning an effective metric classifier from given few-shot training instances. For example, Vinyals et al. \cite{vinyals2016matching} proposed a matching network that classifies query samples by measuring instance-wise consine similarity between queries and supports. Snell et al. \cite{snell2017prototypical} advanced the matching network by further introducing prototypical representations, thereby constructing global category representations for support samples. Santoro et al. \cite{santoro2016meta} proposed a memory-augmented neural network that utilizes stored memory to make query categorization more accurate. Li et al. \cite{li2020boosting} and Zhang et al. \cite{zhang2021prototype} proposed to additionally consider semantic attributes encoded by GloVe\cite{pennington2014glove} or word2vec \cite{mikolov2013efficient}, so as to further improve visual representations of input images. Besides, Xu et al. \cite{xu2022generating} and Yang et al. \cite{yang2023semantic} proposed to further exploit the semantic guidance from CLIP \cite{radford2021learning}, as they found semantic information encoded by CLIP is more effective in learning representative feature embeddings of visual scenes.

\subsection{Incremental Learning}
Incremental learning aims to effectively transfer a model to new categories, while maintaining its previously learned old knowledge as much as possible. Knowledge distillation \cite{hinton2015distilling} has shown its advantages in overcoming a catastrophic forgetting problem \cite{li2017learning}. Aiming to incorporate the knowledge distillation with the data-replay strategy, Rebuffi et al. \cite{rebuffi2017icarl} introduced the exemplar-based knowledge distillation at the cost of extra small storage expenses. Castro et al. \cite{castro2018end} and Kang et al. \cite{kang2022class} pointed out that it is necessary to achieve a good balance between old class knowledge maintenance and new class adaptation. Therefore, the cross-distillation loss and the balanced finetune strategy were utilized in \cite{castro2018end}, while \cite{kang2022class} employed the adaptive feature consolidation strategy to restrict the representation drift of critical old class feature embeddings. Recently, Wang et al. \cite{wang2022foster} advanced the incremental learning model with the gradient boosting strategy, so as to guide the model to effectively learn its residuals to the target one. Liu et al. \cite{liu2021rmm} further enhanced the data-replay strategy by designing the reinforced memory management mechanism. It dynamically adjusts the stored memory information in each incremental step, thereby helping to overcome the sub-optimal memory allocation problem.

\subsection{Incremental Few-shot Semantic Segmentation}
Incremental Few-shot Semantic Segmentation (IFSS), proposed by Cermelli et al. \cite{cermelli2021prototype}, aims at enduing semantic segmentation models with the capability of few-shot incremental learning, thereby making them more suitable to be deployed in open-ended real-world applications. Aiming to address this task, Cermelli et al. \cite{cermelli2021prototype} proposed the prototype-based knowledge distillation. It relieves the catastrophic forgetting issue by constraining the invariance of old class segmentation scores. Moreover, the overfitting to novel categories is suppressed by boosting the consistency between old and updated models. Shi et al. \cite{shi2022incremental} proposed to build hyper-class feature representations, thereby helping to relieve the representation drift during the incremental learning. Furthermore, they adopted a different evaluation protocol than the one employed in \cite{cermelli2021prototype}. Despite the success achieved by these methods, the guidance of visual semantics is ignored in them, which has been proven to play an important role in low-data tasks. Therefore, different from these works, in this paper, we study on how to exploit prior semantic information in IFSS to make segmentation results more accurate.

\begin{figure*}[t!]
  \centering
  \includegraphics[height=8cm]{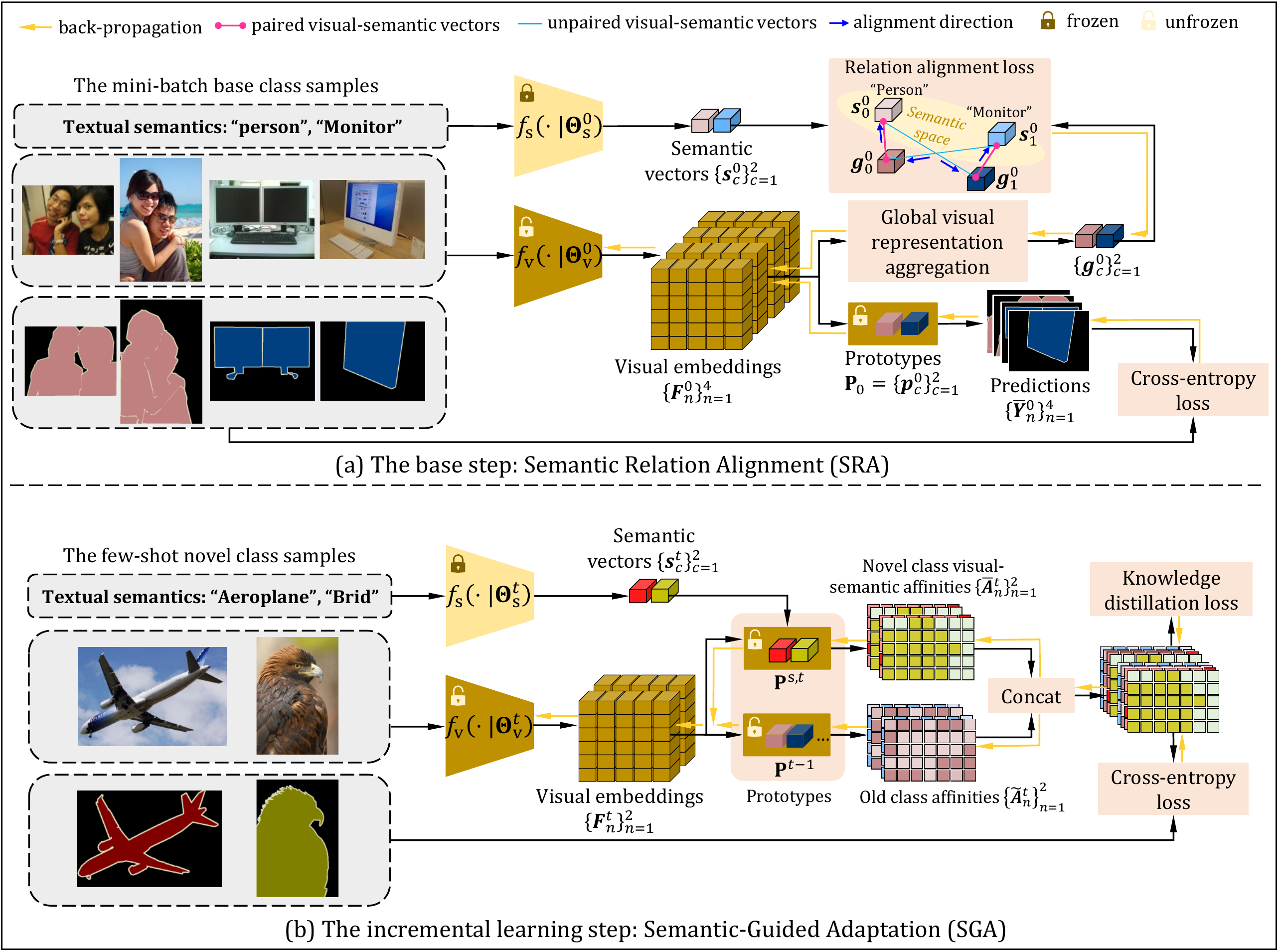}
  \caption{The illustration for our Semantic-guided Relation Alignment and Adaptation (SRAA) method. SRA aims to semantically align the representations of base classes in latent semantic space, and SGA aims at ensuring the visual-semantic affinities of encountered novel categories. Also, ``$f_\mathrm{s}(\cdot|\cdot)$'' and ``$f_\mathrm{v}(\cdot|\cdot)$'' represent a semantic encoder and a visual encoder respectively.}
  \label{fig:2}
\end{figure*}

\section{Methodology}
We elaborate on our proposed method in this section. We first give the preliminaries in Section \ref{sec:prel}. Then, the details about our Semantic-guided Relation Alignment and Adaptation (SRAA) are provided in Section \ref{sec:sraa}.

\subsection{Preliminaries}
\label{sec:prel}
The semantic space of the IFSS model is expanded over time. We define $\bm{\mathrm{C}}^{t}$ as the categories encountered at the step $t$. Therefore, after learning in this step, the semantic space of the model is expanded to $\bm{\mathrm{S}}^{t}=\bm{\mathrm{S}}^{t-1}\bigcup\bm{\mathrm{C}}^{t}$, where $\bm{\mathrm{S}}^{t-1}=\bigcup_{i=0}^{t-1}\bm{\mathrm{C}}^i$ denotes the semantic space learned after the step $t-1$. In each step, the dataset $\bm{\mathrm{D}}^t=\big\{\bm{X}_n^t, \bm{Y}_n^t\big\}_{n=1}^{N_t}$ is provided to update learnable parameters, in which $\bm{X}_n^t$ denotes the $n$-$th$ training image and $\bm{Y}_n^t$ is the label map of $\bm{X}_n^t$. In the IFSS task, the base dataset $\bm{\mathrm{D}}^0$ is provided in the base step (\textit{i.e.}, $t=0$) to initialize the parameters of the model, which contains relatively more training samples. After the base step, the dataset $\bm{\mathrm{D}}^t$ is only in the few-shot setting, i.e., each category contains one or a few labeled training instances, which satisfies the condition $N_t<<N_0$ for $\forall t>=1$. For brevity, in this paper, the categories given in the base step are called base categories, while the categories encountered in the incremental learning stage are termed novel categories. In the step $t$, the model only has access to the dataset $\bm{\mathrm{D}}^t$, and the datasets in the previous steps are unavailable.

\subsection{Semantic-guided Relation Alignment and Adaptation}
\label{sec:sraa}
As described in Figure \ref{fig:2}, our method consists of the two components that are Semantic Relation Alignment (SRA) and Semantic-Guided Adaptation (SGA). These two components are incorporated together to help the model be aware of semantic information of base and novel categories, thereby alleviating the semantic-aliasing issue between them. We elaborate on these two components in the following.

\subsubsection{Semantic Relation Alignment}
\label{sec:sra}
The goal of our SRA is to semantically align base classes in latent semantic space and guide the model to generate semantic-consistent visual representations. We first extract the visual embeddings $\big\{\bm{F}^0_n\big\}_{n=1}^{N_\mathrm{b}}$ from the input images $\big\{\bm{X}_{n}^0\big\}_{n=1}^{N_\mathrm{b}}\subset \bm{\mathrm{D}}^0$ using the visual encoder $f_\mathrm{v}(\cdot|\bm{\Theta}^0_\mathrm{v})$,
\begin{equation}
\big\{\bm{F}^0_n\big\}_{n=1}^{N_\mathrm{b}} = f_\mathrm{v}(\big\{\bm{X}_n^0\big\}_{n=1}^{N_\mathrm{b}}|\bm{\Theta}^0_\mathrm{v})
\label{eq:1}
\end{equation}
where $\bm{\Theta}^0_\mathrm{v}$ indicates the learnable parameters of the visual encoder in the base step, and $N_\mathrm{b}$ denotes the number of images in a mini-batch. Meanwhile, the semantic encoder $f_\mathrm{s}(\cdot|\bm{\Theta}_\mathrm{s}^0)$ encodes the semantic vectors $\big\{\bm{s}^0_c\big\}_{c=1}^{|\bm{\mathrm{C}}_\mathrm{b}|}$ of the categories $\bm{\mathrm{C}}_\mathrm{b}$ involved in the inputs,
\begin{equation}
\big\{\bm{s}^0_c\big\}_{c=1}^{|\bm{\mathrm{C}}_\mathrm{b}|} = f_\mathrm{s}(\bm{\mathrm{C}}_\mathrm{b}|\bm{\Theta}_\mathrm{s}^0)
\label{eq:2}
\end{equation}
in which $\bm{\Theta}_\mathrm{s}^0$ represents the parameters of the semantic encoder, and $\bm{s}^0_c$ denotes the encoded semantic vector about the class $c\in \bm{\mathrm{C}}_\mathrm{b}$. After that, the global visual representations $\big\{\bm{g}^0_c\big\}_{c=1}^{|\bm{\mathrm{C}}_\mathrm{b}|} $ are aggregated for each of the categories $\bm{\mathrm{C}}_\mathrm{b}$,
\begin{equation}
\bm{g}^0_c = \frac{\sum_{n=1}^{N_\mathrm{b}}\sum_{i=1}^{H}\sum_{j=1}^{W}(\bm{F}^0_{n, [i,j]}*\bm{I}^{0,c}_{n, [i,j]})}{\sum_{n=1}^{N_\mathrm{b}}\sum_{i=1}^{H}\sum_{j=1}^{W}\bm{I}^{0,c}_{n, [i,j]}}
\label{eq:3}
\end{equation}
\begin{equation}
s.t.\quad \bm{I}^{0,c}_n=\bm{Y}^0_n==c
\label{eq:4}
\nonumber
\end{equation}
where $\bm{F}^0_{n}\in\mathbb{R}^{H\times W\times D}$ denotes the feature map encoded by the visual encoder $f_\mathrm{v}(\cdot|\bm{\Theta}^0_\mathrm{v})$, and $\bm{I}^{0,c}_{n}\in\mathbb{R}^{H\times W}$ indicates the binary mask about the category $c$. Of note, if the pixel at the position $[i,j]$ of $\bm{X}^0_n$ belongs to the category $c$, $\bm{I}^{0,c}_{n, [i,j]}=1$; otherwise, $\bm{I}^{0,c}_{n, [i,j]}=0$. \par

Aiming to align base class features with their semantics, the relation alignment loss $\mathcal{L}_{\mathrm{align}}$ is employed. It jointly considers the correlations between visual and semantic embeddings that are paired and unpaired,
\begin{align}
  \mathcal{L}_{\mathrm{align}} = \underbrace{\sum_{c_1=1}^{|\bm{\mathrm{C}}_\mathrm{b}|} \sum_{c_2=1, c_2\neq c_1}^{|\bm{\mathrm{C}}_\mathrm{b}|} \frac{\bm{g}^0_{c_1} * \bm{s}^0_{c_2}}{|\bm{\mathrm{C}}_\mathrm{b}|\times (|\bm{\mathrm{C}}_\mathrm{b}|-1)}}_{\text{Unpaired}} - \underbrace{\sum_{c=1}^{|\bm{\mathrm{C}}_\mathrm{b}|} \frac{\bm{g}^0_c * \bm{s}^0_c}{|\bm{\mathrm{C}}_\mathrm{b}|}}_{\text{Paired}}.
  \label{eq:5}
\end{align}
The paired visual-semantic embeddings indicate that the visual vector $\bm{g}_c^0$ and the semantic vector $\bm{s}_c^0$ belong to the same class, and thus $\bm{g}_c^0$ should be aligned to match $\bm{s}_c^0$ in latent space. On the contrary, if the visual embeddings $\bm{g}_{c_1}^0$ and the semantic embeddings $\bm{s}_{c_2}^0$ ($c_2\neq c_1$) are unpaired, the correlations between them should be suppressed to ensure representation discrimination between categories. We minimize the relation alignment loss $\mathcal{L}_{\mathrm{align}}$ w.r.t. the learnable parameters of the visual encoder. Thereby, the model is guided to generate semantic-consistent visual representations, and the base classes embeddings are aligned with their semantic information. \par

Moreover, the segmentation results $\big\{\bar{\bm{Y}}_n^0\in \mathbb{R}^{H\times W\times |\bm{\mathrm{C}}^0|}\big\}_{n=1}^{N_\mathrm{b}}$ with respect to the semantic space $\bm{\mathrm{C}}^0$ are drawn by using the base class prototypical classifier $\bm{\mathrm{P}}^0=\big\{\bm{p}_c^0\big\}_{c=1}^{|\bm{\mathrm{C}}^0|}$, which is shown below:
\begin{align}
\bar{\bm{Y}}_{n, [i,j,c]}^0 & =P(c|\bm{X}_{n, [i,j]}^0,\bm{\mathrm{P}}^0, \bm{\Theta}^0_\mathrm{v})\\
&=\frac{\exp(Sim(\bm{F}_{n,[i,j]}^0,\bm{p}_{c}^0))}{\sum_{c'\in\bm{\mathrm{C}}^0}\exp(Sim(\bm{F}_{n,[i,j]}^0,\bm{p}_{c'}^0))}.\nonumber
\label{eq:7}
\end{align}
In the above equation, $P(c|\bm{X}_{n, [i,j]}^0,\bm{\mathrm{P}}^0, \bm{\Theta}^0_\mathrm{v})$ indicates the probability that the pixel $\bm{X}_{n, [i,j]}^0$ is inferred as the category $c$ according to $\bm{\mathrm{P}}^0$ and $\bm{\Theta}^0_\mathrm{v}$. $Sim(\cdot,\cdot)$ is a similarity metric function, which aims to measures consine similarity between feature embeddings. The cross-entropy loss $\mathcal{L}_{\mathrm{ce}} $ is used to measure the discrepancy between the segmentation results and the groundtruth labels of the inputs,
 \begin{equation}
\mathcal{L}_{\mathrm{ce}} = \frac{1}{N_\mathrm{b}}\sum^{N_\mathrm{b}}_{n=1}CE(\bar{\bm{Y}}_n^0, \bm{Y}_n^0).
\label{eq:8}
\end{equation}
By jointly minimizing the relation alignment loss $\mathcal{L}_{\mathrm{align}}$ and the cross-entropy loss $\mathcal{L}_{\mathrm{ce}}$ during the training process, the model learns to segment base categories while being aware of their semantic information.

\subsubsection{Semantic-Guided Adaptation}
\label{sec:sga}
After the base step, the model is incrementally extended to novel classes. We hope the model can also be aware of the semantic information of encountered novel categories. Therefore, we propose to ensure affinities between visual and semantic embeddings of encountered novel ones. Taking the step $t$ as an example, we first extract the visual embeddings from the images given in the few-shot dataset $\bm{\mathrm{D}}^t$ using the visual encoder $f_\mathrm{v}(\cdot|\bm{\Theta}^t_\mathrm{v})$,
\begin{equation}
\big\{\bm{F}^{t}_n\big\}_{n=1}^{N_{t}} = f_\mathrm{v}(\big\{\bm{X}_n^{t}\big\}_{n=1}^{N_{t}}|\bm{\Theta}^t_\mathrm{v}).
\label{eq:10}
\end{equation}
In the equation, $\bm{\Theta}^t_\mathrm{v}$ indicates the parameters of the visual encoder in the step $t$. Meanwhile, the semantic encoder $f_\mathrm{s}(\cdot|\bm{\Theta}_\mathrm{s}^t)$ encodes the semantic embeddings of the encountered new categories $\bm{\mathrm{C}}^t$,
\begin{equation}
\big\{\bm{s}^{t}_c\big\}_{c=1}^{|\bm{\mathrm{C}}^t|} = f_\mathrm{s}(\bm{\mathrm{C}}^{t}|\bm{\Theta}_\mathrm{s}^t).
\label{eq:11}
\end{equation}
Afterwards, these semantic vectors are used to imprint the weights of the semantic prototypes $\bm{\mathrm{P}}^{\mathrm{s},t}=\big\{\bm{p}^{\mathrm{s},t}_{c}\big\}_{c=1}^{|\bm{\mathrm{C}}^{t}|}$, which are used to guide the finetune process on the novel categories. Specifically, we first calculate the affinities $\big\{\bar{\bm{A}}^{t}_n\in\mathbb{R}^{H\times W\times |\bm{\mathrm{C}}^t|}\big\}_{n=1}^{N_{t}}$ between the visual embeddings of the given images and the semantics of the novel classes according to Eq. \ref{eq:12},
\begin{equation}
\bar{\bm{A}}_{n, [i,j,c]}^{t}=\frac{\bm{F}_{n,[i,j]}^{t} *\bm{p}_{c}^{\mathrm{s},t}}{|\bm{F}_{n,[i,j]}^{t}| *|\bm{p}_{c}^{\mathrm{s},t}|},\,\, s.t., 0<c<=|\bm{\mathrm{C}}^{t}|
\label{eq:12}
\end{equation}
where $\bar{\bm{A}}_{n}^{t}$ denotes the dense visual-semantic affinities about the sample $\bm{X}_n^{t}$, and $\bar{\bm{A}}_{n,[i,j,c]}^{t}$ indicates the affinity between the visual features at the position $[i, j]$ and the semantic embeddings of the category $c\in \bm{\mathrm{C}}^t$. The dense visual-semantic affinities reflect the relation between the visual embeddings and the semantics of the encountered novel classes.

Moreover, the prototypes $\bm{\mathrm{P}}^{t-1}=\big\{\bm{p}_i^{t-1}\big\}_{i=1}^{|\bigcup_{j=0}^{t-1}\bm{\mathrm{C}}^j|}$ learned in the previous steps are utilized to compute the affinities of the current feature maps to the old categories $\big\{\tilde{\bm{A}}^{t}_n\in~\mathbb{R}^{H\times W\times |\bigcup_{j=0}^{t-1}{\bm{\mathrm{C}}^{j}}|}\big\}_{n=1}^{N_{t}}$,
\begin{equation}
\tilde{\bm{A}}_{n, [i,j,c]}^{t}=\frac{\bm{F}_{n,[i,j]}^{t} * \bm{p}_{c}^{t-1}}{|\bm{F}_{n,[i,j]}^{t}| * |\bm{p}_{c}^{t-1}|}, s.t., 0<c<=|\bigcup_{j=0}^{t-1}{\bm{\mathrm{C}}^{j}}|.
\label{eq:13}
\end{equation}
The affinity maps $\tilde{\bm{A}}^{t}_n$ and $\bar{\bm{A}}^{t}_n$ are concatenated together for each sample
\begin{equation}
\bm{A}^{t}_n=\tilde{\bm{A}}^{t}_n \oplus \bar{\bm{A}}^{t}_n,
\label{eq:14}
\end{equation}
thereby producing $\big\{\bm{A}^{t}_n\in~\mathbb{R}^{H\times W\times |\bigcup_{j=0}^{t}{\bm{\mathrm{C}}^{j}}|}\big\}_{n=1}^{N_t}$. We use the cross entropy to constrain the correctness of the affinity maps $\big\{\bm{A}^{t}_n\big\}_{n=1}^{N_t}$
\begin{equation}
\mathcal{L}_{\mathrm{aff}} = \frac{1}{N_t}\sum_{n=1}^{N_t} CE(\bm{A}_n^t, \bm{Y}_n^t),
\label{eq:15}
\end{equation}
so as to guide the visual embeddings of the novel class images to have high correlations to their visual semantics while suppressing the affinities to the old classes. As a result, the feature embeddings of the novel classes can be consistent with their semantic information. Moreover, knowledge distillation is adopted to suppress the overfitting to encountered novel categories
\begin{equation}
\mathcal{L}_{\mathrm{kd}} = \frac{1}{N_t}\sum_{n=1}^{N_t} CE(\bm{A}_n^t, \bm{A}_n^{t-1}),
\label{eq:16}
\end{equation}
where $\bm{A}_n^{t-1}$ denotes the affinities drawn by the model after being trained in the step $t-1$. The joint minimization of $\mathcal{L}_{\mathrm{aff}}$ and $\mathcal{L}_{\mathrm{kd} }$ w.r.t. $\bm{\Theta}^t_\mathrm{s}$, $\bm{\mathrm{P}}^{t-1}$, and $\bm{\mathrm{P}}^{\mathrm{s},t}$ guides the model to be aware of the visual semantics of the encountered novel categories. Meanwhile, the prototypes $\bm{\mathrm{P}}^{t-1}$ and $\bm{\mathrm{P}}^{\mathrm{s},t}$ are optimized to be consistent to reflect the relationships between images and categories, so as to help accurately segment out the objects that belong to the encountered categories from images. After learning in the step $t$, the prototypes are updates: $\bm{\mathrm{P}}^t\leftarrow \hat{\bm{\mathrm{P}}}^{t-1} \bigcup \hat{\bm{\mathrm{P}}}^{\mathrm{s},t}$, where $\hat{\bm{\mathrm{P}}}^{t-1}$ and $\hat{\bm{\mathrm{P}}}^{\mathrm{s},t}$ indicate the prototypes $\bm{\mathrm{\mathrm{P}}}^{t-1}$ and $\bm{\mathrm{P}}^{\mathrm{s},t}$ after being optimized in the current step. The updated prototypes and visual encoder are employed to draw segmentation results for all the encountered classes, as same as the process shown in Eq. 5.

\section{Experiments}
Experiments are provided in this section to validate our proposed method. We first introduce the datasets in Section 4.1 and give our implementation details in Section 4.2. Then, we report the main experimental results in Section 4.3, and conduct the ablation study in Section 4.4.

\subsection{Datasets}
We evaluate the proposed method on the two public semantic segmentation datasets that are PASCAL VOC 2012 \cite{everingham2015pascal,hariharan2011semantic} and COCO \cite{caesar2018coco,lin2014microsoft}. The PASCAL VOC 2012 dataset consists of 10582 training images and 1449 test images, which are collected from 20 different categories. Following the previous work \cite{cermelli2021prototype}, we divide these 20 categories into four folds and each fold includes five categories, which is summarized in Table \ref{tab:voc}. In addition, on the COCO dataset, 80 categories are used to evaluate the performance of the model, including about 110k training samples and 5k test samples. As presented in Table \ref{tab:coco}, the 80 categories of this dataset are split into four folds as well, which is the same as \cite{cermelli2021prototype} does. For the cross validation on both these datasets, we use the categories of three folds to build the base set, while the categories of the rest one fold are used for testing.

\begin{table}
  \footnotesize
  \caption{The dataset split on the PASCAL VOC 2012 dataset.}
  \label{tab:voc}
  \begin{tabular}{c|c}
    \midrule
    Split & Categories \\
    \midrule
    5-0 & \makecell[l]{aeroplane, bicycle, bird, boat, bottle} \\
    \midrule
    5-1 & \makecell[l]{bus, car, cat, chair, cow} \\
    \midrule
    5-2 & \makecell[l]{table, dog, horse, motorbike, person} \\
    \midrule
    5-3 & \makecell[l]{plant, sheep, sofa, train, tv-monitor} \\
    \midrule
  \end{tabular}
\end{table}

\begin{table}
  \footnotesize
  \caption{The dataset split on the COCO dataset.}
  \label{tab:coco}
  \begin{tabular}{c|c}
    \midrule
    Split & Categories \\
    \midrule
    20-0 & \makecell[l]{person, airplane, boat, parking meter, dog, elephant, refrigerator,\\ backpack, suitcase, sports ball, skateboard, wine glass, spoon,\\ sandwich, hot dog, chair, dining table, mouse, microwave,  scissors} \\
    \midrule
    20-1 & \makecell[l]{bicycle, bus, traffic light, bench, horse, bear, umbrella, frisbee,\\ kite, surfboard, cup, bowl, orange, pizza,couch, toilet, remote,\\ oven, book, teddy bear} \\
    \midrule
    20-2 & \makecell[l]{car, train, fire hydrant, bird, sheep, zebra, handbag, skis,\\ baseball bat, tennis racket, fork, banana, broccoli, donut,\\ potted plant, tv, keyboard,toaster, clock, hair drier}\\
    \midrule
    20-3 & \makecell[l]{motorcycle, truck, stop sign, cat, cow, giraffe, tie, snowboard,\\ baseball glove, bottle, knife, apple, carrot, cake, bed, laptop,\\ cell phone, sink, vase, toothbrush} \\
    \midrule
  \end{tabular}
\end{table}

\begin{table*}
\footnotesize
\caption{The experimental results on the PASCAL VOC 2012 dataset. In the table, ``FT'' indicates directly finetuning the model on novel classes using the cross-entropy loss, and ``HM'' indicates the harmonic mean of the mIoU on base and novel classes. Also, the first-place and the second-place result in each column are highlighted in bold font and underscore respectively.}
    \begin{subtable}[t]{0.495\linewidth}
      \centering
        \caption{The results under the single few-shot step setting.}
        \setlength{\tabcolsep}{1.1mm}{        
        \begin{tabular}{c|cc|c|cc|c|cc|c}
            \toprule
             &  \multicolumn{3}{c|}{1-shot} &  \multicolumn{3}{c|}{2-shot} & \multicolumn{3}{c}{5-shot} \\
             \midrule
             & \multicolumn{2}{c|}{mIoU (\%)} & & \multicolumn{2}{c|}{mIoU (\%)} & & \multicolumn{2}{c|}{mIoU (\%)} & \\
             \cline{2-3}\cline{5-6}\cline{8-9}
            Method & base & novel & HM & base & novel & HM & base & novel & HM \\
            \midrule
            FT & 58.3 & 9.7 & 16.7 & 59.1 & 19.7 & 29.5 & 55.8 & 29.6 & 38.7 \\
            WI \cite{nichol2018first} & 62.7 & 15.5 & 24.8 & 63.3 & 19.2 & 29.5 & 63.3 & 21.7 & 32.3 \\
            DWI \cite{gidaris2018dynamic} & \underbar{64.3} & 15.4 & 24.8 & \textbf{64.8} & 19.8 & 30.4 & \underline{64.9} & 23.5 & 34.5 \\
            RT \cite{tian2020rethinking} & 59.1 & 12.1 & 20.1 & 60.9 & 21.6 & 31.9 & 60.4 & 27.5 & 37.8 \\
            AMP \cite{siam2019adaptive} & 57.5 & 16.7 & 25.8 & 54.4 & 18.8 & 27.9 & 51.9 & 18.9 & 27.7 \\
            SPN \cite{xian2019semantic} & 59.8 & 16.3 & 25.6 & 60.8 & 26.3 & 36.7 & 58.4 & \underline{33.4} & 42.5 \\
            LWF \cite{li2017learning} & 61.5 & 10.7 & 18.2 & 63.6 & 18.9 & 29.2 & 59.7 & 30.9 & 40.8 \\
            ILT \cite{michieli2019incremental} & \underline{64.3} & 13.6 & 22.5 & \underline{64.2} & 23.1 & 34.0 & 61.4 & 32.0 & 42.1 \\
            MIB \cite{cermelli2020modeling} & 61.0 & 5.2 & 9.7 & 63.5 & 12.7 & 21.1 & \textbf{65.0} & 28.1 & 39.3 \\
            PIFS \cite{cermelli2021prototype} & 60.9 & \underline{18.6} & \underline{28.4} & 60.5 & \underline{26.4} & \underline{36.8} & 60.0 & \underline{33.4} & \underline{42.8} \\
            \midrule
            Ours & \textbf{65.2} & \textbf{19.1} & \textbf{29.5} & 62.7 & \textbf{27.4} & \textbf{38.1} & 63.8 & \textbf{36.7} & \textbf{46.6} \\
            \bottomrule
        \end{tabular}}
    \end{subtable}
    \begin{subtable}[t]{0.495\linewidth}
          \centering
          \caption{The results under the multiple few-shot step setting.}
          \setlength{\tabcolsep}{1.1mm}{
          \begin{tabular}{c|cc|c|cc|c|cc|c}
            \toprule
             &  \multicolumn{3}{c|}{1-shot} &  \multicolumn{3}{c|}{2-shot} & \multicolumn{3}{c}{5-shot} \\
            \midrule
            & \multicolumn{2}{c|}{mIoU (\%)} & & \multicolumn{2}{c|}{mIoU (\%)} & & \multicolumn{2}{c|}{mIoU (\%)} & \\
            \cline{2-3}\cline{5-6}\cline{8-9}
            Method & base & novel & HM & base & novel & HM & base & novel & HM \\
            \midrule
            FT & 47.2 & 3.9 & 7.2 & 53.5 & 4.4 & 8.1 & 58.7 & 7.7 & 13.6 \\
            WI \cite{nichol2018first} & \underline{66.6} & 16.1 & 25.9 & \underline{66.6} & 19.8 & 30.5 & \underline{66.6} & 21.9 & 33.0 \\
            DWI \cite{gidaris2018dynamic} & \textbf{67.2} & 16.3 & 26.2 & \textbf{67.5} & 21.6 & 32.7 & \textbf{67.6} & 25.4 & 36.9 \\
            RT \cite{tian2020rethinking} & 49.2 & 5.8 & 10.4 & 36.0 & 4.9 & 8.6 & 45.1 & 10.0 & 16.4 \\
            AMP \cite{siam2019adaptive} & 58.6 & 14.5 & 23.2 & 58.4 & 16.3 & 25.5 & 57.1 & 17.2 & 26.4 \\
            SPN \cite{xian2019semantic} & 49.8 & 8.1 & 13.9 & 56.4 & 10.4 & 17.6 & 61.6 & 16.3 & 25.8 \\
            LWF \cite{li2017learning} & 42.1 & 3.3 & 6.2 & 51.6 & 3.9 & 7.3 & 59.8 & 7.5 & 13.4 \\
            ILT \cite{michieli2019incremental} & 43.7 & 3.3 & 6.1 & 52.2 & 4.4 & 8.1 & 59.0 & 7.9 & 13.9 \\
            MIB \cite{cermelli2020modeling} & 43.9 & 2.6 & 4.9 & 51.9 & 2.1 & 4.0 & 60.9 & 5.8 & 10.5 \\
            PIFS \cite{cermelli2021prototype} & 64.1 & \underline{16.9} & \underline{26.7} & 65.2 & \underline{23.7} & \underline{34.8} & 64.5 & \underline{27.5} & \underline{38.6} \\
            \midrule
            Ours & 66.4 & \textbf{18.8} & \textbf{29.3} & 65.1 & \textbf{26.4} & \textbf{37.6} & 64.3 & \textbf{28.7} & \textbf{39.7} \\
            \bottomrule
        \end{tabular}}
    \end{subtable}
    \label{voc-r}
\end{table*}

\begin{table*}
\footnotesize
\centering
\caption{The experimental results on the COCO dataset.}
    \begin{subtable}[t]{0.495\linewidth}
        \centering
        \caption{The results under the single few-shot step setting.}
        \setlength{\tabcolsep}{1.1mm}{
        \begin{tabular}{c|cc|c|cc|c|cc|c}
            \toprule
             &  \multicolumn{3}{c|}{1-shot} &  \multicolumn{3}{c|}{2-shot} & \multicolumn{3}{c}{5-shot} \\
            \midrule
            & \multicolumn{2}{c|}{mIoU (\%)} & & \multicolumn{2}{c|}{mIoU (\%)} & & \multicolumn{2}{c|}{mIoU (\%)} & \\
            \cline{2-3}\cline{5-6}\cline{8-9}
             Method & base & novel & HM & base & novel & HM & base & novel & HM \\
            \midrule
            FT & 41.2 & 4.1 & 7.5 & 41.5 & 7.3 & 12.4 & 41.6 & 12.3 & 19.0 \\
            WI \cite{nichol2018first} &  43.8 & 6.9 & 11.9 & 44.2 & 7.9 & 13.5 & 43.6 & 8.7 & 14.6 \\
            DWI \cite{gidaris2018dynamic} & \underline{44.5} & 7.5 & 12.8 & 45.0 & 9.4 & 15.6 & 44.9 & 12.1 & 19.1 \\
            RT \cite{tian2020rethinking} & \textbf{46.2} & 5.8 & 10.2 & \textbf{46.7} & 8.8 & 14.8 & \underline{46.9} & 13.7 & 21.2 \\
            AMP \cite{siam2019adaptive}  & 37.5 & 7.4 & 12.4 & 35.7 & 8.8 & 14.2 & 34.6 & 11.0 & 16.7 \\
            SPN \cite{xian2019semantic} & 43.5 & 6.7 & 11.7 & 43.7 & 10.2 & 16.5 & 43.7 & 15.6 & 22.9 \\
            LWF \cite{li2017learning}  & 43.9 & 3.8 & 7.0 & 44.3 & 7.1 & 12.3 & 44.6 & 12.9 & 20.1 \\
            ILT \cite{michieli2019incremental} & \textbf{46.2} & 4.4 & 8.0 & \underline{46.3} & 6.5 & 11.5 & \textbf{47.0} & 11.0 & 17.8 \\
            MIB \cite{cermelli2020modeling} & 43.8 & 3.5 & 6.5 & 44.4 & 6.0 & 10.6 & 44.7 & 11.9 & 18.8 \\
            PIFS \cite{cermelli2021prototype} & 40.8 & \underline{8.2} & \underline{13.7} & 40.9 & \underline{11.1} & \underline{17.5} & 42.8 & \underline{15.7} & \underline{23.0} \\
            \midrule
            Ours & 41.2 & \textbf{9.3} & \textbf{15.2} & 42.1 & \textbf{12.7} & \textbf{19.5} & 42.6 & \textbf{17.1} & \textbf{24.4} \\
            \bottomrule
        \end{tabular}}
    \end{subtable}
    \begin{subtable}[t]{0.495\linewidth}
        \centering
         \caption{The results under the multiple few-shot step setting.}
         \setlength{\tabcolsep}{1.1mm}{
          \begin{tabular}{c|cc|c|cc|c|cc|c}
            \toprule
             &  \multicolumn{3}{c|}{1-shot} &  \multicolumn{3}{c|}{2-shot} & \multicolumn{3}{c}{5-shot} \\
            \midrule
             & \multicolumn{2}{c|}{mIoU (\%)} & & \multicolumn{2}{c|}{mIoU (\%)} & & \multicolumn{2}{c|}{mIoU (\%)} & \\
             \cline{2-3}\cline{5-6}\cline{8-9}
            Method & base & novel & HM & base & novel & HM & base & novel & HM \\
            \midrule
            FT & 38.5 & 4.8 & 8.5 & 40.3 & 6.8 & 11.6 & 39.5 & 11.5 & 17.8 \\
            WI \cite{nichol2018first} & \textbf{46.3} & 8.3 & 14.1 & \underline{46.5} & 9.3 & 15.5 & \underline{46.3} & 10.3 & 16.9 \\
            DWI \cite{gidaris2018dynamic} & \underline{46.2} & 9.2 & 15.3 & \underline{46.5} & 11.4 & 18.3 & \textbf{46.6} & 14.5 & 22.1 \\
            RT \cite{tian2020rethinking} &  38.4 & 5.2 & 9.2 & 43.8 & 10.1 & 16.4 & 44.1 & 16.0 & 23.5 \\
            AMP \cite{siam2019adaptive} & 36.6 & 7.9 & 13.0 & 36.0 & 9.2 & 14.7 & 33.2 & 11.0 & 16.5 \\
            SPN \cite{xian2019semantic} & 40.3 & 8.7 & 14.3 & 41.7 & 12.5 & 19.2 & 41.4 & 18.2 & \underline{25.3} \\
            LWF \cite{li2017learning} & 41.0 & 4.1 & 7.5 & 42.7 & 6.5 & 11.3 & 42.3 & 12.6 & 19.4 \\
            ILT \cite{michieli2019incremental} & 43.7 & 6.2 & 10.9 & \textbf{47.1} & 10.0 & 16.5 & 45.3 & 15.3 & 22.9 \\
            MIB \cite{cermelli2020modeling} & 40.4 & 3.1 & 5.8 & 42.7 & 5.2 & 9.3 & 43.8 & 11.5 & 18.2 \\
            PIFS \cite{cermelli2021prototype} & 40.4 & \underline{10.4} & \underline{16.5} & 40.1 & \underline{13.1} & \underline{19.8} & 41.1 & \underline{18.3} & \underline{25.3} \\
            \midrule
            Ours & 40.7 & \textbf{11.3} & \textbf{17.7} & 40.5 & \textbf{15.2} & \textbf{22.1} & 41.0 & \textbf{19.7} & \textbf{26.6} \\
            \bottomrule
        \end{tabular}}
    \end{subtable}
    \label{tab:coco-r}
\end{table*}

\subsection{Implementation Details}
Our codes are implemented using PyTorch and run on the tesla V100 GPU card. In the experiments, the SGD optimizer is adopted to optimize the learnable parameters of our model based on the poly learning rate. For the experiments on the PASCAL VOC 2012 and the COCO dataset, the training details are slightly different. Specifically, on the PASCAL VOC 2012 dataset, we set the number of the epochs as 30 on the base step and 400 during the incremental learning. Also, in each phase, the initial learning rate of the optimizer is set as 0.01. On the COCO dataset, we train the model on the base set for 50 epochs with the initial poly learning rate 0.02.  Moreover, during the incremental learning stage, the epochs are set as 400, and the initial learning rate is set as 0.01. Following the previous work \cite{cermelli2021prototype}, we evaluate our method in both the single few-shot step setting and the multiple few-shot step setting based on the cross validation protocol. The single few-shot step setting indicates that all novel categories are given at once in an incremental step, while the multiple few-shot step setting means novel categories are progressively given in multiple steps. We employ the mean Intersection-over-Union (mIoU) metric in our experiments to evaluate the performance of the method. Besides, we build our semantic encoder using CLIP, due to its powerful capability in encoding semantic information \cite{xu2022generating,yang2023semantic}. Following the previous methods \cite{xu2022generating,yang2023semantic}, we freeze the parameters of the semantic encoder during the training process, i.e., $\bm{\Theta}^t_\mathrm{s}=\bm{\Theta}^0_\mathrm{s}$ for $\forall t>=1$. Meanwhile, as same as \cite{cermelli2021prototype}, we build our visual encoder by using resnet101~\cite{he2016deep}.

\begin{figure*}[t!]
  \centering
  \includegraphics[height=5.2cm]{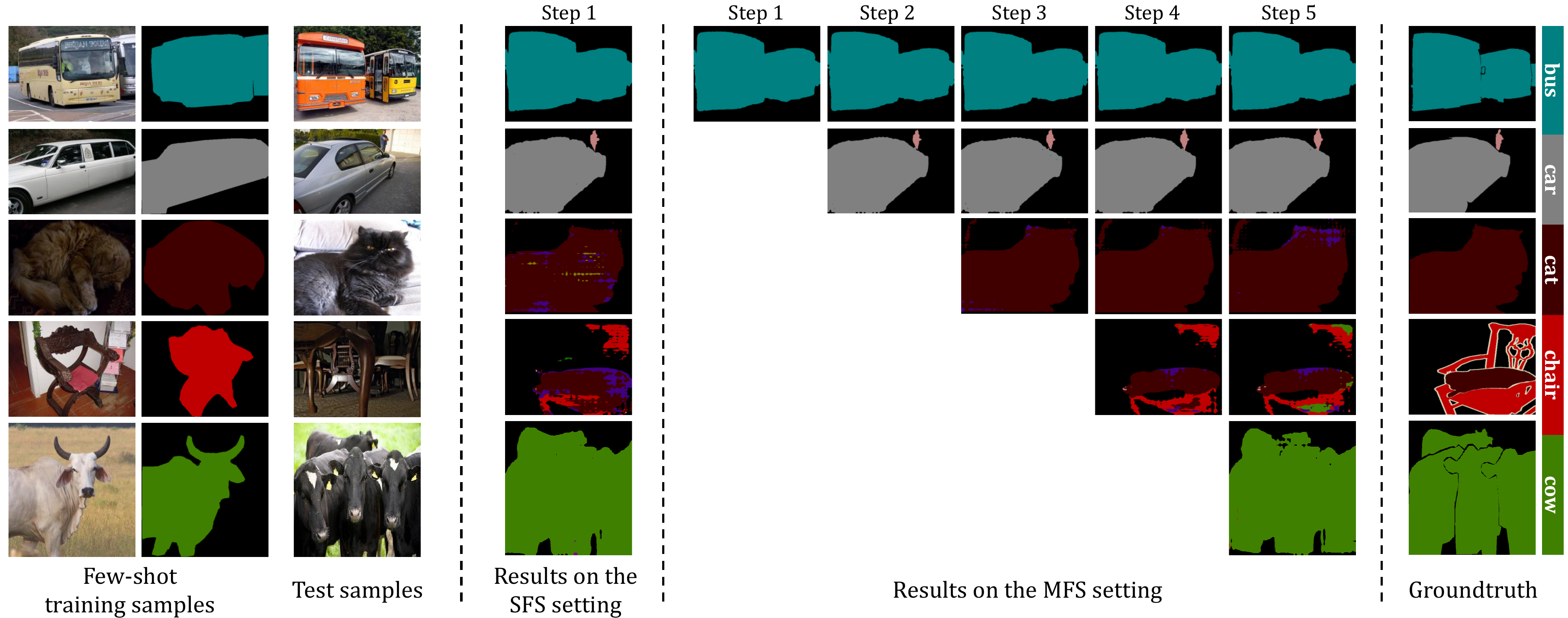}
  \caption{The visualization for the step-by-step segmentation results of our method on both the Single Few-shot Step (SFS) setting and the Multiple Few-shot Step (MFS) setting according to a labeled sample per category. The model is progressively extended to the novel classes in the MFS setting. In the SFS setting, the novel classes are given at once in an incremental step.}
  \label{fig:6}
\end{figure*}

\begin{figure}[t!]
  \centering
  \includegraphics[height=3cm]{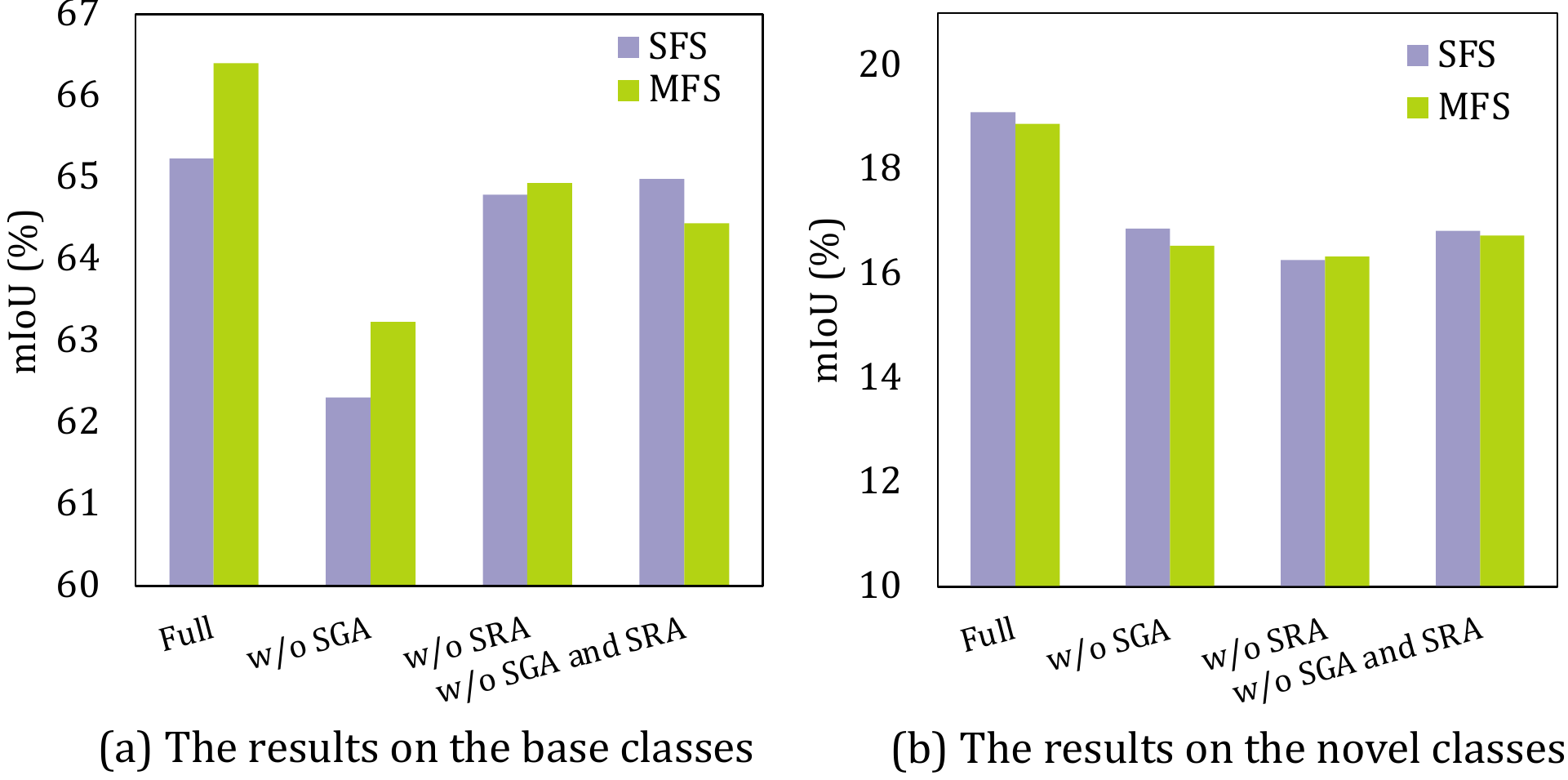}
  \caption{The ablation study for our method on both the Single Few-shot Step (SFS) setting and the Multiple Few-shot Step (MFS) setting, which is conducted on the 1-shot task of the PASCAL VOC 2012 dataset.}
  \label{fig:a} 
\end{figure}

\begin{figure}[t!]
  \centering
  \includegraphics[height=4.2cm]{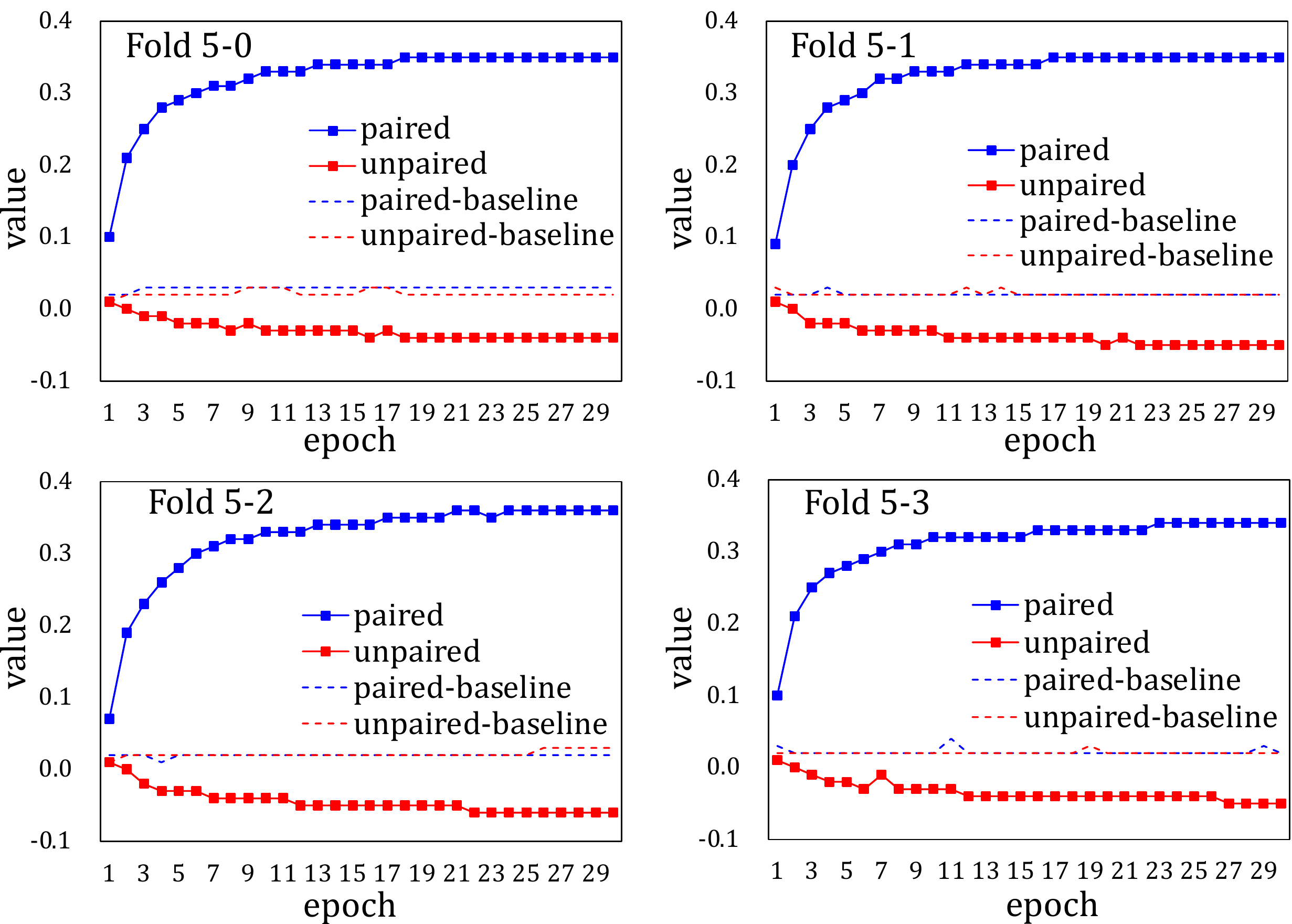}
  \caption{The values of the paired and the unpaired term in our relation alignment loss $\mathcal{L}_{\mathrm{align}}$ during the training process. To better reflect value change, the baseline curves are also drawn in the figure, which reflect the values of these two terms when $\mathcal{L}_{\mathrm{align}}$ is not employed during the training. The above experiments are conducted on the PASCAL VOC 2012 dataset.}
  \label{fig:3}
\end{figure}

\begin{figure}[t!]
  \centering
  \includegraphics[height=3.2cm]{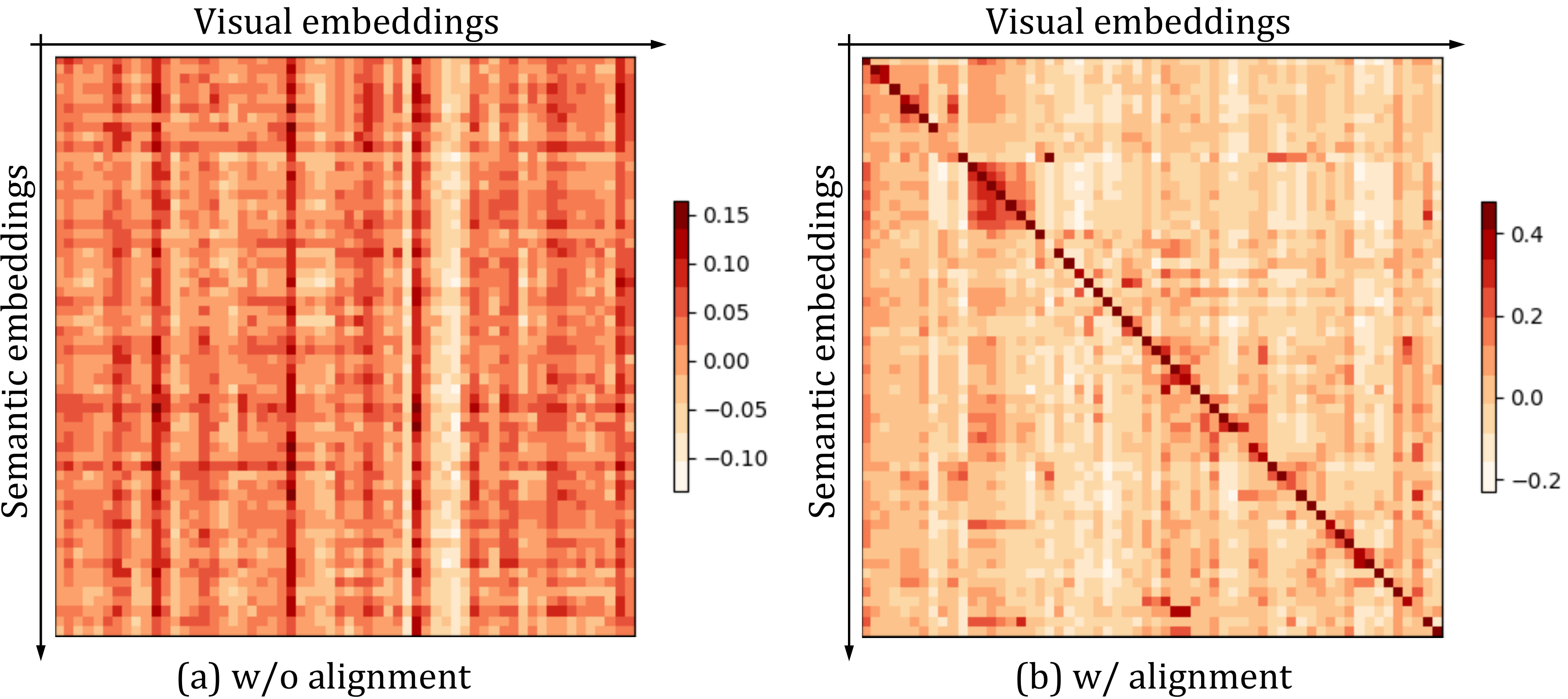}
  \caption{The mean affinities between the visual and the semantic embeddings with or without being aligned by our SRA, which is conducted on the COCO dataset. The values in the diagonal are the affinities between the visual and the semantic embeddings that belong to the same class, while the others are the affinities between them that are unpaired.}
  \label{fig:4}
\end{figure}

\begin{figure}[t!]
  \centering
  \includegraphics[height=7.2cm]{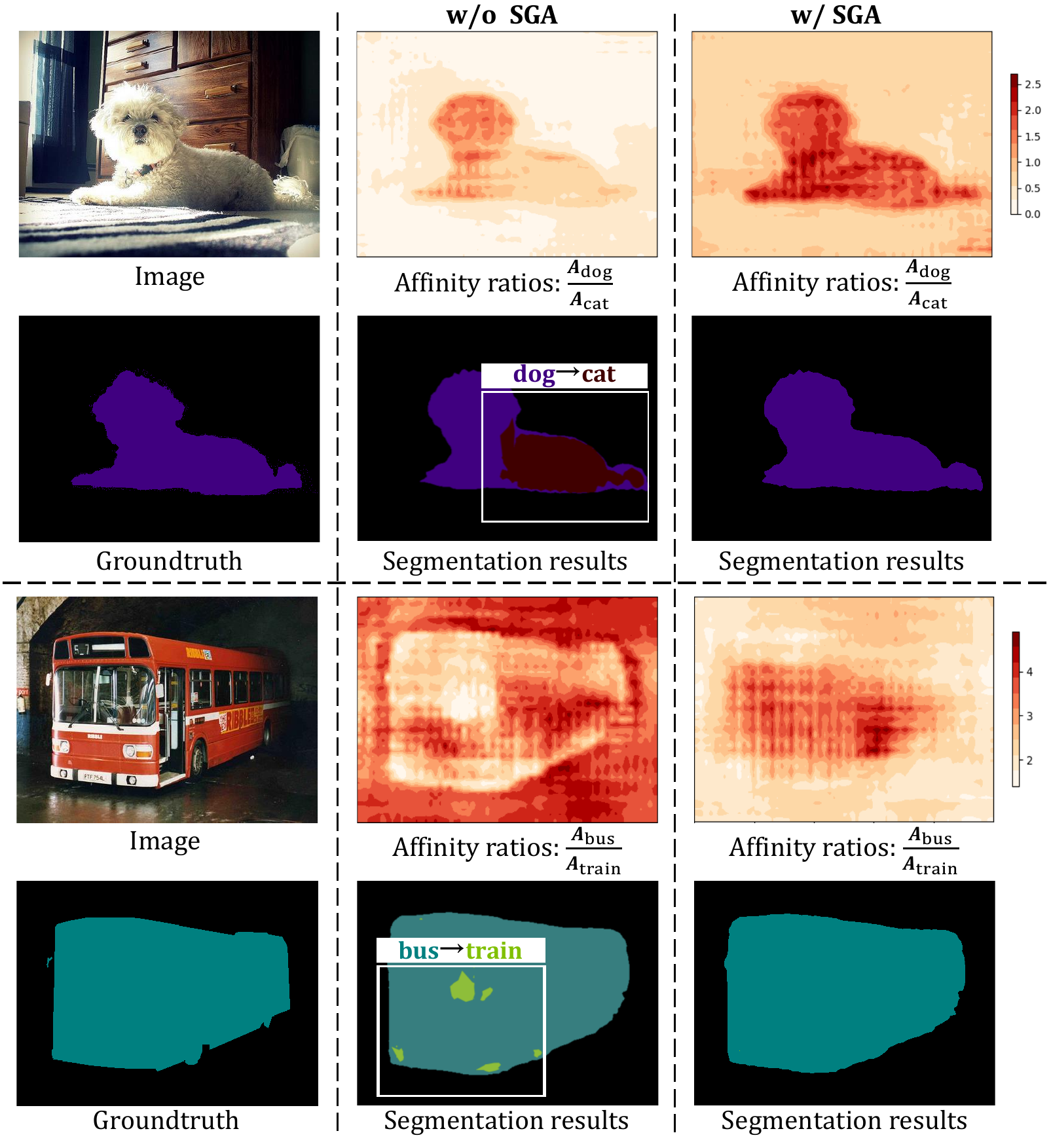}
  \caption{The visualized analysis for the influence of our SGA. In the figure, $\bm{A}_{\mathrm{dog}}$, $\bm{A}_{\mathrm{cat}}$, $\bm{A}_{\mathrm{bus}}$, or $\bm{A}_{\mathrm{train}}$ denotes the affinity map of an image to the category ``dog'', ``cat'', ``bus'', or ``train''. $\frac{\bm{A}_{\mathrm{dog}}}{\bm{A}_{\mathrm{cat}}}$ or $\frac{\bm{A}_{\mathrm{bus}}}{\bm{A}_{\mathrm{train}}}$ indicates the ratio map between $\bm{A}_{\mathrm{dog}}$ and $\bm{A}_{\mathrm{cat}}$ or $\bm{A}_{\mathrm{bus}}$ and $\bm{A}_{\mathrm{train}}$}
  \label{fig:c}
\end{figure}

\subsection{Main Results}
The results of our method on the PASCAL VOC 2012 and the COCO dataset are summarized in Table \ref{voc-r} and Table \ref{tab:coco-r} respectively. According to these results, we have the following observations. On the PASCAL VOC 2012 dataset, our method achieves higher mIoU on both base and novel categories than that of FT, RT, AMP, SPN, and PIFS under the single few-shot step setting. Despite the performance of MIB, ILT, LWF, DWI, and WI on base categories is better than that of ours in some cases, our method obviously achieves higher mIoU on novel categories. For example, on the 2-shot task, the novel class mIoU of our method is $14.7 \%$, $4.3 \%$, $8.5 \%$, $7.6\%$, and $8.2\%$ higher than that of MIB, ILT, LWF, DWI, and WI respectively. In the meantime, our method shows its superiority on the PASCAL VOC 2012 dataset under the multiple few-shot step setting as well. For example, on all the 1-shot, the 2-shot, and the 5-shot task, the novel class mIoU of PIFS is lower than that of our proposed method. Similar results can also be found on the experiments of the COCO dataset. For example, in the single few-shot step setting, the performance of our method is better than that of PIFS and AMP on both base and novel categories. Although the base class mIoU of our method is lower than that of the several compared methods, it consistently shows higher mIoU on encountered novel categories, e.g., our method's novel class mIoU is $5.8\%$, $4.9\%$, $5.5\%$, $2.6\%$, $3.5\%$, $1.8\%$, and $2.4\%$ higher than that of MIB, ILT, LWF, SPN, RT, DWI, and WI on the 1-shot task. Moreover, under the multiple few-shot step setting, the novel class mIoU of our proposed method is higher than that of all the compared methods in the table. For example, our method's novel class mIoU is $0.9\%$, $2.1\%$, and $1.4\%$ higher than that of the second-place method PIFS on the 1-shot, the 2-shot, and the 5-shot task. In Figure \ref{fig:6}, we give our step-by-step segmentation results for the encountered novel categories under the two different settings. The results indicate that according to only one training instance per novel category, our method can still achieve the promising semantic segmentation results. Moreover, when encountering new classes, it can still maintain high effectiveness in segmenting the categories learned in the previous few-shot learning steps.

\subsection{Ablation Study}
In this subsection, we first study the influence of SGA and SRA on accuracy in Figure~\ref{fig:a}. ``w/o SGA'' indicates that the semantic guidance is not considered during the adaptation procedure. Thus, the prototypes about novel categories are imprinted by the mean visual embeddings of given samples. ``w/o SRA'' indicates that SRA is not employed, and base class embeddings are not aligned with their semantics. On one hand, the cooperation of our SRA and SGA (i.e., ``Full'') can achieve higher mIoU than that of the baseline model (``w/o SGA and SRA'') on both base and novel classes under the two different settings, which demonstrates that the appropriate use of prior semantic information can make segmentation results more accurate. On the other hand, the removal of SGA or SRA (i.e., ``w/o SGA'' or ``w/o SRA'') leads to an obvious performance drop, thereby validating the importance of these two components. The results also indicate that semantic information should be considered in both the base and the incremental learning stage. Otherwise, the training inconsistency between phases will reduce segmentation accuracy. \par

Then, in Figure \ref{fig:3}, we visualize the values of the paired and the unpaired term in the relation alignment loss $\mathcal{L}_{\mathrm{align}}$ during the training process. The results indicate that $\mathcal{L}_{\mathrm{align}}$ can be optimized stably. On one hand, with the epoch increases, the paired term is maximized progressively, thereby constraining that visual and semantic embeddings belonging to the same category have relatively high correlations. On the other hand, the minimization of the unpaired term suppresses the similarity between visual and semantic embeddings that are unpaired. As a result, the visual embeddings belonging to the same class are guided to have high semantic correlations, while the semantic correlations of different categories are limited. We also visualize the mean affinities between the visual and the semantic embeddings that are aligned by our method. The results in Figure~\ref{fig:4} validate the effectiveness of our SRA again, e.g., SRA can obviously rectify visual embeddings to better match their semantic information. 

\begin{figure}[t!]
  \centering
  \includegraphics[height=8.6cm]{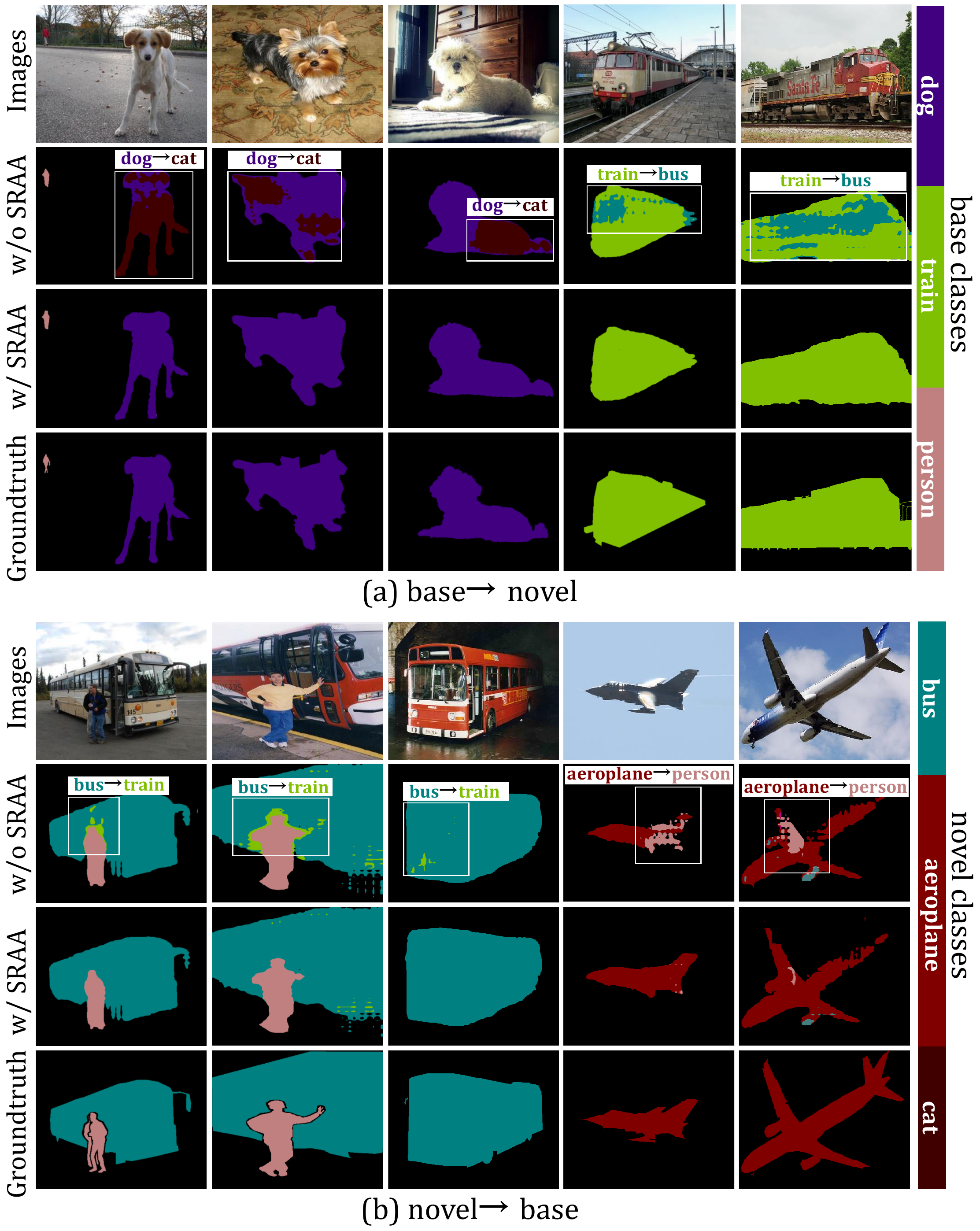}
  \caption{The visualized analysis for the influence of our method on alleviating the semantic-aliasing issue in both the ``base$\rightarrow$novel'' and the ``novel$\rightarrow$base'' scenario, which is conducted on the PASCAL VOC 2012 dataset under the 1-shot setting. Notice that ``A$\rightarrow$B'' indicates regions belonging to A are incorrectly segmented as B.} 
  \label{fig:5}
\end{figure}

The analysis for the influence of our SGA is provided in Figure~\ref{fig:c}. As can seen from the figure, without SGA, the sample about ``dog'' is incorrectly segmented as the class ``cat'' due to the incorrect affinity information, e.g., the affinity ratios $\frac{\bm{A}_{\mathrm{dog}}}{\bm{A}_{\mathrm{cat}}}$ have low values in the target regions. In contrast, the use of SGA can obviously boost the affinities to the target class, while suppressing the affinities to ``cat''. In this way, the segmentation results can be more accurate. For the instance ``bus'', the affinity ratios $\frac{\bm{A}_{\mathrm{bus}}}{\bm{A}_{\mathrm{train}}}$ show low values for a part of the target regions when not employing our SGA. Moreover, in the background areas, the affinity ratios have the incorrect high values. By leveraging the guidance of visual semantics, our method can rectify these incorrect affinities, thereby making segmentation results more accurate. Finally, in Figure \ref{fig:5}, we also provide the qualitative analysis for the influence of our SRAA method on final segmentation results. The experimental results consistently validate the superiority of our method as well. 

\section{Conclusion}
In this paper, we propose to alleviate the semantic-aliasing issue in IFSS by conducting Semantic-guided Relation Alignment and Adaptation (SRAA). On one hand, we introduce Semantic Relation Alignment (SRA) in the base step, aiming to semantically align the representations of base categories and guide the model to generate semantic-consistent feature representations. On the other hand, we employ Semantic-Guided Adaptation (SGA) to incrementally adapt the model to novel classes. It ensures the visual-semantic affinities of encountered novel categories, so as to make their feature embeddings be consistent with the corresponding semantic information. By considering the semantic information of both the base and the novel categories, the semantic-aliasing issue can be relieved. Currently, it is still very challenging to incrementally achieve accurate segmentation results for objects with complex and varied boundaries in IFSS. In the future, we plan to overcome this problem by fully considering the fine-grained information of local features.

\bibliographystyle{ACM-Reference-Format}
\bibliography{acmart} 


\begin{thebibliography}{42}


\ifx \showCODEN    \undefined \def \showCODEN     #1{\unskip}     \fi
\ifx \showDOI      \undefined \def \showDOI       #1{#1}\fi
\ifx \showISBNx    \undefined \def \showISBNx     #1{\unskip}     \fi
\ifx \showISBNxiii \undefined \def \showISBNxiii  #1{\unskip}     \fi
\ifx \showISSN     \undefined \def \showISSN      #1{\unskip}     \fi
\ifx \showLCCN     \undefined \def \showLCCN      #1{\unskip}     \fi
\ifx \shownote     \undefined \def \shownote      #1{#1}          \fi
\ifx \showarticletitle \undefined \def \showarticletitle #1{#1}   \fi
\ifx \showURL      \undefined \def \showURL       {\relax}        \fi
\providecommand\bibfield[2]{#2}
\providecommand\bibinfo[2]{#2}
\providecommand\natexlab[1]{#1}
\providecommand\showeprint[2][]{arXiv:#2}

\bibitem[Caesar et~al\mbox{.}(2018)]%
        {caesar2018coco}
\bibfield{author}{\bibinfo{person}{Holger Caesar}, \bibinfo{person}{Jasper Uijlings}, {and} \bibinfo{person}{Vittorio Ferrari}.} \bibinfo{year}{2018}\natexlab{}.
\newblock \showarticletitle{Coco-stuff: Thing and stuff classes in context}. In \bibinfo{booktitle}{\emph{Proceedings of the Conference on Computer Vision and Pattern Recognition}}. \bibinfo{pages}{1209--1218}.
\newblock


\bibitem[Castro et~al\mbox{.}(2018)]%
        {castro2018end}
\bibfield{author}{\bibinfo{person}{Francisco~M Castro}, \bibinfo{person}{Manuel~J Mar{\'\i}n-Jim{\'e}nez}, \bibinfo{person}{Nicol{\'a}s Guil}, \bibinfo{person}{Cordelia Schmid}, {and} \bibinfo{person}{Karteek Alahari}.} \bibinfo{year}{2018}\natexlab{}.
\newblock \showarticletitle{End-to-end incremental learning}. In \bibinfo{booktitle}{\emph{Proceedings of the European Conference on Computer Vision}}. \bibinfo{pages}{233--248}.
\newblock


\bibitem[Cermelli et~al\mbox{.}(2020)]%
        {cermelli2020modeling}
\bibfield{author}{\bibinfo{person}{Fabio Cermelli}, \bibinfo{person}{Massimiliano Mancini}, \bibinfo{person}{Samuel~Rota Bulo}, \bibinfo{person}{Elisa Ricci}, {and} \bibinfo{person}{Barbara Caputo}.} \bibinfo{year}{2020}\natexlab{}.
\newblock \showarticletitle{Modeling the background for incremental learning in semantic segmentation}. In \bibinfo{booktitle}{\emph{Proceedings of the Conference on Computer Vision and Pattern Recognition}}. \bibinfo{pages}{9233--9242}.
\newblock


\bibitem[Cermelli et~al\mbox{.}(2021)]%
        {cermelli2021prototype}
\bibfield{author}{\bibinfo{person}{Fabio Cermelli}, \bibinfo{person}{Massimiliano Mancini}, \bibinfo{person}{Yongqin Xian}, \bibinfo{person}{Zeynep Akata}, {and} \bibinfo{person}{Barbara Caputo}.} \bibinfo{year}{2021}\natexlab{}.
\newblock \showarticletitle{Prototype-based Incremental Few-Shot Semantic Segmentation}. In \bibinfo{booktitle}{\emph{Proceedings of the British Machine Vision Conference}}. BMVC 2021, \bibinfo{pages}{484}.
\newblock


\bibitem[Chen et~al\mbox{.}(2017a)]%
        {chen2017deeplab}
\bibfield{author}{\bibinfo{person}{Liang-Chieh Chen}, \bibinfo{person}{George Papandreou}, \bibinfo{person}{Iasonas Kokkinos}, \bibinfo{person}{Kevin Murphy}, {and} \bibinfo{person}{Alan~L Yuille}.} \bibinfo{year}{2017}\natexlab{a}.
\newblock \showarticletitle{Deeplab: Semantic image segmentation with deep convolutional nets, atrous convolution, and fully connected crfs}.
\newblock \bibinfo{journal}{\emph{IEEE Transactions on Pattern Analysis and Machine Intelligence}} \bibinfo{volume}{40}, \bibinfo{number}{4} (\bibinfo{year}{2017}), \bibinfo{pages}{834--848}.
\newblock


\bibitem[Chen et~al\mbox{.}(2017b)]%
        {chen2017rethinking}
\bibfield{author}{\bibinfo{person}{Liang-Chieh Chen}, \bibinfo{person}{George Papandreou}, \bibinfo{person}{Florian Schroff}, {and} \bibinfo{person}{Hartwig Adam}.} \bibinfo{year}{2017}\natexlab{b}.
\newblock \showarticletitle{Rethinking atrous convolution for semantic image segmentation}.
\newblock \bibinfo{journal}{\emph{arXiv preprint arXiv:1706.05587}} (\bibinfo{year}{2017}).
\newblock


\bibitem[Dosovitskiy et~al\mbox{.}(2020)]%
        {dosovitskiy2020image}
\bibfield{author}{\bibinfo{person}{Alexey Dosovitskiy}, \bibinfo{person}{Lucas Beyer}, \bibinfo{person}{Alexander Kolesnikov}, \bibinfo{person}{Dirk Weissenborn}, \bibinfo{person}{Xiaohua Zhai}, \bibinfo{person}{Thomas Unterthiner}, \bibinfo{person}{Mostafa Dehghani}, \bibinfo{person}{Matthias Minderer}, \bibinfo{person}{Georg Heigold}, \bibinfo{person}{Sylvain Gelly}, {et~al\mbox{.}}} \bibinfo{year}{2020}\natexlab{}.
\newblock \showarticletitle{An image is worth 16x16 words: Transformers for image recognition at scale}.
\newblock \bibinfo{journal}{\emph{arXiv preprint arXiv:2010.11929}} (\bibinfo{year}{2020}).
\newblock


\bibitem[Everingham et~al\mbox{.}(2015)]%
        {everingham2015pascal}
\bibfield{author}{\bibinfo{person}{Mark Everingham}, \bibinfo{person}{SM~Ali Eslami}, \bibinfo{person}{Luc Van~Gool}, \bibinfo{person}{Christopher~KI Williams}, \bibinfo{person}{John Winn}, {and} \bibinfo{person}{Andrew Zisserman}.} \bibinfo{year}{2015}\natexlab{}.
\newblock \showarticletitle{The pascal visual object classes challenge: A retrospective}.
\newblock \bibinfo{journal}{\emph{International Journal of Computer Vision}}  \bibinfo{volume}{111} (\bibinfo{year}{2015}), \bibinfo{pages}{98--136}.
\newblock


\bibitem[Gidaris and Komodakis(2018)]%
        {gidaris2018dynamic}
\bibfield{author}{\bibinfo{person}{Spyros Gidaris} {and} \bibinfo{person}{Nikos Komodakis}.} \bibinfo{year}{2018}\natexlab{}.
\newblock \showarticletitle{Dynamic few-shot visual learning without forgetting}. In \bibinfo{booktitle}{\emph{Proceedings of the Conference on Computer Vision and Pattern Recognition}}. \bibinfo{pages}{4367--4375}.
\newblock


\bibitem[Hariharan et~al\mbox{.}(2011)]%
        {hariharan2011semantic}
\bibfield{author}{\bibinfo{person}{Bharath Hariharan}, \bibinfo{person}{Pablo Arbel{\'a}ez}, \bibinfo{person}{Lubomir Bourdev}, \bibinfo{person}{Subhransu Maji}, {and} \bibinfo{person}{Jitendra Malik}.} \bibinfo{year}{2011}\natexlab{}.
\newblock \showarticletitle{Semantic contours from inverse detectors}. In \bibinfo{booktitle}{\emph{Proceedings of the International Conference on Computer Vision}}. IEEE, \bibinfo{pages}{991--998}.
\newblock


\bibitem[He et~al\mbox{.}(2016)]%
        {he2016deep}
\bibfield{author}{\bibinfo{person}{Kaiming He}, \bibinfo{person}{Xiangyu Zhang}, \bibinfo{person}{Shaoqing Ren}, {and} \bibinfo{person}{Jian Sun}.} \bibinfo{year}{2016}\natexlab{}.
\newblock \showarticletitle{Deep residual learning for image recognition}. In \bibinfo{booktitle}{\emph{Proceedings of the IEEE conference on computer vision and pattern recognition}}. \bibinfo{pages}{770--778}.
\newblock


\bibitem[Hinton et~al\mbox{.}(2015)]%
        {hinton2015distilling}
\bibfield{author}{\bibinfo{person}{Geoffrey Hinton}, \bibinfo{person}{Oriol Vinyals}, {and} \bibinfo{person}{Jeff Dean}.} \bibinfo{year}{2015}\natexlab{}.
\newblock \showarticletitle{Distilling the knowledge in a neural network}.
\newblock \bibinfo{journal}{\emph{arXiv preprint arXiv:1503.02531}} (\bibinfo{year}{2015}).
\newblock


\bibitem[Huang et~al\mbox{.}(2019)]%
        {huang2019ccnet}
\bibfield{author}{\bibinfo{person}{Zilong Huang}, \bibinfo{person}{Xinggang Wang}, \bibinfo{person}{Lichao Huang}, \bibinfo{person}{Chang Huang}, \bibinfo{person}{Yunchao Wei}, {and} \bibinfo{person}{Wenyu Liu}.} \bibinfo{year}{2019}\natexlab{}.
\newblock \showarticletitle{Ccnet: Criss-cross attention for semantic segmentation}. In \bibinfo{booktitle}{\emph{Proceedings of the International Conference on Computer Vision}}. \bibinfo{pages}{603--612}.
\newblock


\bibitem[Kang et~al\mbox{.}(2022)]%
        {kang2022class}
\bibfield{author}{\bibinfo{person}{Minsoo Kang}, \bibinfo{person}{Jaeyoo Park}, {and} \bibinfo{person}{Bohyung Han}.} \bibinfo{year}{2022}\natexlab{}.
\newblock \showarticletitle{Class-incremental learning by knowledge distillation with adaptive feature consolidation}. In \bibinfo{booktitle}{\emph{Proceedings of the Conference on Computer Vision and Pattern Recognition}}. \bibinfo{pages}{16071--16080}.
\newblock


\bibitem[Li et~al\mbox{.}(2020)]%
        {li2020boosting}
\bibfield{author}{\bibinfo{person}{Aoxue Li}, \bibinfo{person}{Weiran Huang}, \bibinfo{person}{Xu Lan}, \bibinfo{person}{Jiashi Feng}, \bibinfo{person}{Zhenguo Li}, {and} \bibinfo{person}{Liwei Wang}.} \bibinfo{year}{2020}\natexlab{}.
\newblock \showarticletitle{Boosting few-shot learning with adaptive margin loss}. In \bibinfo{booktitle}{\emph{Proceedings of the Conference on Computer Vision and Pattern Recognition}}. \bibinfo{pages}{12576--12584}.
\newblock


\bibitem[Li and Hoiem(2017)]%
        {li2017learning}
\bibfield{author}{\bibinfo{person}{Zhizhong Li} {and} \bibinfo{person}{Derek Hoiem}.} \bibinfo{year}{2017}\natexlab{}.
\newblock \showarticletitle{Learning without forgetting}.
\newblock \bibinfo{journal}{\emph{IEEE Transactions on Pattern Analysis and Machine Intelligence}} \bibinfo{volume}{40}, \bibinfo{number}{12} (\bibinfo{year}{2017}), \bibinfo{pages}{2935--2947}.
\newblock


\bibitem[Lin et~al\mbox{.}(2014)]%
        {lin2014microsoft}
\bibfield{author}{\bibinfo{person}{Tsung-Yi Lin}, \bibinfo{person}{Michael Maire}, \bibinfo{person}{Serge Belongie}, \bibinfo{person}{James Hays}, \bibinfo{person}{Pietro Perona}, \bibinfo{person}{Deva Ramanan}, \bibinfo{person}{Piotr Doll{\'a}r}, {and} \bibinfo{person}{C~Lawrence Zitnick}.} \bibinfo{year}{2014}\natexlab{}.
\newblock \showarticletitle{Microsoft coco: Common objects in context}. In \bibinfo{booktitle}{\emph{Proceedings of the European Conference on Computer Vision}}. Springer, \bibinfo{pages}{740--755}.
\newblock


\bibitem[Liu et~al\mbox{.}(2021)]%
        {liu2021rmm}
\bibfield{author}{\bibinfo{person}{Yaoyao Liu}, \bibinfo{person}{Bernt Schiele}, {and} \bibinfo{person}{Qianru Sun}.} \bibinfo{year}{2021}\natexlab{}.
\newblock \showarticletitle{RMM: Reinforced memory management for class-incremental learning}.
\newblock \bibinfo{journal}{\emph{Advances in Neural Information Processing Systems}}  \bibinfo{volume}{34} (\bibinfo{year}{2021}), \bibinfo{pages}{3478--3490}.
\newblock


\bibitem[Long et~al\mbox{.}(2015)]%
        {long2015fully}
\bibfield{author}{\bibinfo{person}{Jonathan Long}, \bibinfo{person}{Evan Shelhamer}, {and} \bibinfo{person}{Trevor Darrell}.} \bibinfo{year}{2015}\natexlab{}.
\newblock \showarticletitle{Fully convolutional networks for semantic segmentation}. In \bibinfo{booktitle}{\emph{Proceedings of the Conference on Computer Vision and Pattern Recognition}}. \bibinfo{pages}{3431--3440}.
\newblock


\bibitem[Michieli and Zanuttigh(2019)]%
        {michieli2019incremental}
\bibfield{author}{\bibinfo{person}{Umberto Michieli} {and} \bibinfo{person}{Pietro Zanuttigh}.} \bibinfo{year}{2019}\natexlab{}.
\newblock \showarticletitle{Incremental learning techniques for semantic segmentation}. In \bibinfo{booktitle}{\emph{Proceedings of the International Conference on Computer Vision Workshops}}. \bibinfo{pages}{0--0}.
\newblock


\bibitem[Mikolov et~al\mbox{.}(2013)]%
        {mikolov2013efficient}
\bibfield{author}{\bibinfo{person}{Tomas Mikolov}, \bibinfo{person}{Kai Chen}, \bibinfo{person}{Greg Corrado}, {and} \bibinfo{person}{Jeffrey Dean}.} \bibinfo{year}{2013}\natexlab{}.
\newblock \showarticletitle{Efficient estimation of word representations in vector space}.
\newblock \bibinfo{journal}{\emph{arXiv preprint arXiv:1301.3781}} (\bibinfo{year}{2013}).
\newblock


\bibitem[Nichol et~al\mbox{.}(2018)]%
        {nichol2018first}
\bibfield{author}{\bibinfo{person}{Alex Nichol}, \bibinfo{person}{Joshua Achiam}, {and} \bibinfo{person}{John Schulman}.} \bibinfo{year}{2018}\natexlab{}.
\newblock \showarticletitle{On first-order meta-learning algorithms}.
\newblock \bibinfo{journal}{\emph{arXiv preprint arXiv:1803.02999}} (\bibinfo{year}{2018}).
\newblock


\bibitem[Pennington et~al\mbox{.}(2014)]%
        {pennington2014glove}
\bibfield{author}{\bibinfo{person}{Jeffrey Pennington}, \bibinfo{person}{Richard Socher}, {and} \bibinfo{person}{Christopher~D Manning}.} \bibinfo{year}{2014}\natexlab{}.
\newblock \showarticletitle{Glove: Global vectors for word representation}. In \bibinfo{booktitle}{\emph{Proceedings of the Conference on Empirical Methods in Natural Language Processing}}. \bibinfo{pages}{1532--1543}.
\newblock


\bibitem[Radford et~al\mbox{.}(2021)]%
        {radford2021learning}
\bibfield{author}{\bibinfo{person}{Alec Radford}, \bibinfo{person}{Jong~Wook Kim}, \bibinfo{person}{Chris Hallacy}, \bibinfo{person}{Aditya Ramesh}, \bibinfo{person}{Gabriel Goh}, \bibinfo{person}{Sandhini Agarwal}, \bibinfo{person}{Girish Sastry}, \bibinfo{person}{Amanda Askell}, \bibinfo{person}{Pamela Mishkin}, \bibinfo{person}{Jack Clark}, {et~al\mbox{.}}} \bibinfo{year}{2021}\natexlab{}.
\newblock \showarticletitle{Learning transferable visual models from natural language supervision}. In \bibinfo{booktitle}{\emph{Proceedings of the International Conference on Machine Learning}}. PMLR, \bibinfo{pages}{8748--8763}.
\newblock


\bibitem[Rebuffi et~al\mbox{.}(2017)]%
        {rebuffi2017icarl}
\bibfield{author}{\bibinfo{person}{Sylvestre-Alvise Rebuffi}, \bibinfo{person}{Alexander Kolesnikov}, \bibinfo{person}{Georg Sperl}, {and} \bibinfo{person}{Christoph~H Lampert}.} \bibinfo{year}{2017}\natexlab{}.
\newblock \showarticletitle{icarl: Incremental classifier and representation learning}. In \bibinfo{booktitle}{\emph{Proceedings of the Conference on Computer Vision and Pattern Recognition}}. \bibinfo{pages}{2001--2010}.
\newblock


\bibitem[Santoro et~al\mbox{.}(2016)]%
        {santoro2016meta}
\bibfield{author}{\bibinfo{person}{Adam Santoro}, \bibinfo{person}{Sergey Bartunov}, \bibinfo{person}{Matthew Botvinick}, \bibinfo{person}{Daan Wierstra}, {and} \bibinfo{person}{Timothy Lillicrap}.} \bibinfo{year}{2016}\natexlab{}.
\newblock \showarticletitle{Meta-learning with memory-augmented neural networks}. In \bibinfo{booktitle}{\emph{Proceedings of the International Conference on Machine Learning}}. PMLR, \bibinfo{pages}{1842--1850}.
\newblock


\bibitem[Shi et~al\mbox{.}(2022)]%
        {shi2022incremental}
\bibfield{author}{\bibinfo{person}{Guangchen Shi}, \bibinfo{person}{Yirui Wu}, \bibinfo{person}{Jun Liu}, \bibinfo{person}{Shaohua Wan}, \bibinfo{person}{Wenhai Wang}, {and} \bibinfo{person}{Tong Lu}.} \bibinfo{year}{2022}\natexlab{}.
\newblock \showarticletitle{Incremental few-shot semantic segmentation via embedding adaptive-update and hyper-class representation}. In \bibinfo{booktitle}{\emph{Proceedings of the ACM International Conference on Multimedia}}. \bibinfo{pages}{5547--5556}.
\newblock


\bibitem[Siam et~al\mbox{.}(2019)]%
        {siam2019adaptive}
\bibfield{author}{\bibinfo{person}{Mennatullah Siam}, \bibinfo{person}{Boris Oreshkin}, {and} \bibinfo{person}{Martin Jagersand}.} \bibinfo{year}{2019}\natexlab{}.
\newblock \showarticletitle{Adaptive masked proxies for few-shot segmentation}.
\newblock \bibinfo{journal}{\emph{arXiv preprint arXiv:1902.11123}} (\bibinfo{year}{2019}).
\newblock


\bibitem[Snell et~al\mbox{.}(2017)]%
        {snell2017prototypical}
\bibfield{author}{\bibinfo{person}{Jake Snell}, \bibinfo{person}{Kevin Swersky}, {and} \bibinfo{person}{Richard Zemel}.} \bibinfo{year}{2017}\natexlab{}.
\newblock \showarticletitle{Prototypical networks for few-shot learning}.
\newblock \bibinfo{journal}{\emph{Advances in Neural Information Processing Systems}}  \bibinfo{volume}{30} (\bibinfo{year}{2017}).
\newblock


\bibitem[Strudel et~al\mbox{.}(2021)]%
        {strudel2021segmenter}
\bibfield{author}{\bibinfo{person}{Robin Strudel}, \bibinfo{person}{Ricardo Garcia}, \bibinfo{person}{Ivan Laptev}, {and} \bibinfo{person}{Cordelia Schmid}.} \bibinfo{year}{2021}\natexlab{}.
\newblock \showarticletitle{Segmenter: Transformer for semantic segmentation}. In \bibinfo{booktitle}{\emph{Proceedings of the International Conference on Computer Vision}}. \bibinfo{pages}{7262--7272}.
\newblock


\bibitem[Tian et~al\mbox{.}(2020)]%
        {tian2020rethinking}
\bibfield{author}{\bibinfo{person}{Yonglong Tian}, \bibinfo{person}{Yue Wang}, \bibinfo{person}{Dilip Krishnan}, \bibinfo{person}{Joshua~B Tenenbaum}, {and} \bibinfo{person}{Phillip Isola}.} \bibinfo{year}{2020}\natexlab{}.
\newblock \showarticletitle{Rethinking few-shot image classification: a good embedding is all you need?}. In \bibinfo{booktitle}{\emph{Proceedings of the European Conference on Computer Vision}}. Springer, \bibinfo{pages}{266--282}.
\newblock


\bibitem[Vaswani et~al\mbox{.}(2017)]%
        {vaswani2017attention}
\bibfield{author}{\bibinfo{person}{Ashish Vaswani}, \bibinfo{person}{Noam Shazeer}, \bibinfo{person}{Niki Parmar}, \bibinfo{person}{Jakob Uszkoreit}, \bibinfo{person}{Llion Jones}, \bibinfo{person}{Aidan~N Gomez}, \bibinfo{person}{{\L}ukasz Kaiser}, {and} \bibinfo{person}{Illia Polosukhin}.} \bibinfo{year}{2017}\natexlab{}.
\newblock \showarticletitle{Attention is all you need}.
\newblock \bibinfo{journal}{\emph{Advances in Neural Information Processing Systems}}  \bibinfo{volume}{30} (\bibinfo{year}{2017}).
\newblock


\bibitem[Vinyals et~al\mbox{.}(2016)]%
        {vinyals2016matching}
\bibfield{author}{\bibinfo{person}{Oriol Vinyals}, \bibinfo{person}{Charles Blundell}, \bibinfo{person}{Timothy Lillicrap}, \bibinfo{person}{Daan Wierstra}, {et~al\mbox{.}}} \bibinfo{year}{2016}\natexlab{}.
\newblock \showarticletitle{Matching networks for one shot learning}.
\newblock \bibinfo{journal}{\emph{Advances in Neural Information Processing Systems}}  \bibinfo{volume}{29} (\bibinfo{year}{2016}).
\newblock


\bibitem[Wang et~al\mbox{.}(2022)]%
        {wang2022foster}
\bibfield{author}{\bibinfo{person}{Fu-Yun Wang}, \bibinfo{person}{Da-Wei Zhou}, \bibinfo{person}{Han-Jia Ye}, {and} \bibinfo{person}{De-Chuan Zhan}.} \bibinfo{year}{2022}\natexlab{}.
\newblock \showarticletitle{Foster: Feature boosting and compression for class-incremental learning}. In \bibinfo{booktitle}{\emph{Proceedings of the European Conference on Computer Vision}}. Springer, \bibinfo{pages}{398--414}.
\newblock


\bibitem[Xian et~al\mbox{.}(2019)]%
        {xian2019semantic}
\bibfield{author}{\bibinfo{person}{Yongqin Xian}, \bibinfo{person}{Subhabrata Choudhury}, \bibinfo{person}{Yang He}, \bibinfo{person}{Bernt Schiele}, {and} \bibinfo{person}{Zeynep Akata}.} \bibinfo{year}{2019}\natexlab{}.
\newblock \showarticletitle{Semantic projection network for zero-and few-label semantic segmentation}. In \bibinfo{booktitle}{\emph{Proceedings of the Conference on Computer Vision and Pattern Recognition}}. \bibinfo{pages}{8256--8265}.
\newblock


\bibitem[Xie et~al\mbox{.}(2021)]%
        {xie2021segformer}
\bibfield{author}{\bibinfo{person}{Enze Xie}, \bibinfo{person}{Wenhai Wang}, \bibinfo{person}{Zhiding Yu}, \bibinfo{person}{Anima Anandkumar}, \bibinfo{person}{Jose~M Alvarez}, {and} \bibinfo{person}{Ping Luo}.} \bibinfo{year}{2021}\natexlab{}.
\newblock \showarticletitle{SegFormer: Simple and efficient design for semantic segmentation with transformers}.
\newblock \bibinfo{journal}{\emph{Advances in Neural Information Processing Systems}}  \bibinfo{volume}{34} (\bibinfo{year}{2021}), \bibinfo{pages}{12077--12090}.
\newblock


\bibitem[Xu and Le(2022)]%
        {xu2022generating}
\bibfield{author}{\bibinfo{person}{Jingyi Xu} {and} \bibinfo{person}{Hieu Le}.} \bibinfo{year}{2022}\natexlab{}.
\newblock \showarticletitle{Generating representative samples for few-shot classification}. In \bibinfo{booktitle}{\emph{Proceedings of the Conference on Computer Vision and Pattern Recognition}}. \bibinfo{pages}{9003--9013}.
\newblock


\bibitem[Yang et~al\mbox{.}(2023)]%
        {yang2023semantic}
\bibfield{author}{\bibinfo{person}{Fengyuan Yang}, \bibinfo{person}{Ruiping Wang}, {and} \bibinfo{person}{Xilin Chen}.} \bibinfo{year}{2023}\natexlab{}.
\newblock \showarticletitle{Semantic Guided Latent Parts Embedding for Few-Shot Learning}. In \bibinfo{booktitle}{\emph{Proceedings of the Winter Conference on Applications of Computer Vision}}. \bibinfo{pages}{5447--5457}.
\newblock


\bibitem[Zhang et~al\mbox{.}(2021)]%
        {zhang2021prototype}
\bibfield{author}{\bibinfo{person}{Baoquan Zhang}, \bibinfo{person}{Xutao Li}, \bibinfo{person}{Yunming Ye}, \bibinfo{person}{Zhichao Huang}, {and} \bibinfo{person}{Lisai Zhang}.} \bibinfo{year}{2021}\natexlab{}.
\newblock \showarticletitle{Prototype completion with primitive knowledge for few-shot learning}. In \bibinfo{booktitle}{\emph{Proceedings of the Conference on Computer Vision and Pattern Recognition}}. \bibinfo{pages}{3754--3762}.
\newblock


\bibitem[Zhang et~al\mbox{.}(2018)]%
        {zhang2018context}
\bibfield{author}{\bibinfo{person}{Hang Zhang}, \bibinfo{person}{Kristin Dana}, \bibinfo{person}{Jianping Shi}, \bibinfo{person}{Zhongyue Zhang}, \bibinfo{person}{Xiaogang Wang}, \bibinfo{person}{Ambrish Tyagi}, {and} \bibinfo{person}{Amit Agrawal}.} \bibinfo{year}{2018}\natexlab{}.
\newblock \showarticletitle{Context encoding for semantic segmentation}. In \bibinfo{booktitle}{\emph{Proceedings of the Conference on Computer Vision and Pattern Recognition}}. \bibinfo{pages}{7151--7160}.
\newblock


\bibitem[Zhao et~al\mbox{.}(2017)]%
        {zhao2017pyramid}
\bibfield{author}{\bibinfo{person}{Hengshuang Zhao}, \bibinfo{person}{Jianping Shi}, \bibinfo{person}{Xiaojuan Qi}, \bibinfo{person}{Xiaogang Wang}, {and} \bibinfo{person}{Jiaya Jia}.} \bibinfo{year}{2017}\natexlab{}.
\newblock \showarticletitle{Pyramid scene parsing network}. In \bibinfo{booktitle}{\emph{Proceedings of the Conference on Computer Vision and Pattern Recognition}}. \bibinfo{pages}{2881--2890}.
\newblock


\bibitem[Zheng et~al\mbox{.}(2021)]%
        {zheng2021rethinking}
\bibfield{author}{\bibinfo{person}{Sixiao Zheng}, \bibinfo{person}{Jiachen Lu}, \bibinfo{person}{Hengshuang Zhao}, \bibinfo{person}{Xiatian Zhu}, \bibinfo{person}{Zekun Luo}, \bibinfo{person}{Yabiao Wang}, \bibinfo{person}{Yanwei Fu}, \bibinfo{person}{Jianfeng Feng}, \bibinfo{person}{Tao Xiang}, \bibinfo{person}{Philip~HS Torr}, {et~al\mbox{.}}} \bibinfo{year}{2021}\natexlab{}.
\newblock \showarticletitle{Rethinking semantic segmentation from a sequence-to-sequence perspective with transformers}. In \bibinfo{booktitle}{\emph{Proceedings of the Conference on Computer Vision and Pattern Recognition}}. \bibinfo{pages}{6881--6890}.
\newblock


\end{thebibliography}
\end{document}